%% file: article1.tex
\documentclass[11pt,pdftex]{article}
\usepackage[a4paper, total={6.8in, 8in}]{geometry}

\usepackage[utf8]{inputenc}
\usepackage{amsmath}
\usepackage{amsfonts}
\usepackage{amssymb}
\usepackage{cite}
\usepackage{graphicx}
\graphicspath{{Figs/}}
\input{macros_gpi.tex}

\title{Deep Learning and Inverse Problems}

\author{
Ali Mohammad-Djafari $^{1,2}$ orcid number:{0000-0003-0678-7759}, \\ 
Ning Chu$^{2,3}$, Li Wang$^{4}$, Liang Yu$^{5,6}$ 
\\ ~ \\ 
{\small \btabu{l}
$^1$ \quad International Science Consulting and Training (ISCT), 91440 Bures sur Yvette, France; djafari@ieee.org
\\
$^2$ \quad Zhejiang Shangfeng special blower company, Shaoxing 312352, China; chuning1983@sina.com
\\
$^3$ \quad Mechanical and Electrical Eng. Coll., Hainan Vocational Univ of Science and Tech. Haikou 571126, China
\\
$^4$ \quad Central South University, Changsha, China; li.wang.csu@csu.edu.cn
\\ 
$^5$ \quad Shanghai Jiao Tong Univ., Shaghai 200240, China; Liang.Yu@sjtu.edu.cn 
\\
$^6$ \quad Northwestern Polytechnical Univ., Xian 710072, China. 
\\ ~\\ 
Presented at MaxEnt 2023:   
International Workshop on Bayesian Inference and Maximum Entropy \\ 
Methods in Science and Engineering, 
Max Planck Institut, Garching, Germany, July 3-7, 2023.
\\ 
A modified and combined version of this paper will appear in MaxEnt2023 Proceedings.
\etabu
}}

\date{}

\begin{document}

\maketitle

\begin{abstract}
Machine Learning (ML) methods and tools have gained great success in many data, signal, image and video processing tasks, such as classification, clustering, object detection, semantic segmentation, language processing, Human-Machine interface, etc. In computer vision, image and video processing, these methods are mainly based on Neural Networks (NN) and in particular Convolutional NN (CNN), and more generally Deep NN. 
\\ 
Inverse problems arise anywhere we have indirect measurement. As, in general those inverse problems are ill-posed, to obtain satisfactory solutions for them needs prior information. Different regularization methods have been proposed where the problem becomes the optimization of a criterion with a likelihood term and a regularization term. The main difficulty however, in great dimensional real applications, remains the computational cost. Using NN, and in particular Deep Learning (DL) surrogate models and approximate computation can become very helpful.
\\ 
In this work, we focus on NN and DL particularly adapted for inverse problems. We consider two cases: First the case where the forward operator is known and used as physics constraint, the second more general data driven DL methods.  
\\ ~\\ 
{\bf key words: } Neural Network, Deep Learning (DL), Inverse problems, Physics based DL
\end{abstract}

\section{Introduction}
In science and any engineering problem, we need to observe (measure) quantities. Some quantities are directly observable (e.g.; Length), and some others are not (e.g.; Temperature). For example, to measure the temperature, we need an instrument (thermometer) that measures the length of the liquid in the thermometer tube, which can be related to the temperature. We may also wants to observe its variation in time or its spatial distribution. One way to measure the spatial distribution of the temperature is using an Infra-Red (IR) camera. 
But, in general, all these instruments, give indirect measurements, related to what we really want to measure through some mathematical relation, called Forward model. Then, we have to infer on the desired unknown from the observed data, using this forward model or a surrogate one \cite{7043317}. 
 
As, in general, many inverse problems are ill-posed, many classical methods for finding well-posed solutions for them are mainly based on regularization theory. We may mention those, in particular,  which are based on the optimization of a criterion with two parts: 
a data-model output matching criterion and a regularization term. Different criteria for these two terms and a great number of standard and advanced optimization algorithms have been proposed and used with great success.  
When these two terms are distances, they can have a Bayesian Maximum A Posteriori (MAP) interpretation where these two terms correspond, respectively, to the likelihood and prior probability models \cite{7043317}.
 
The Bayesian approach gives more flexibility in choosing these terms via the likelihood and the prior probability distributions. This flexibility goes much farther with the hierarchical models and appropriate hidden variables  \cite{5445046}. 
Also, the possibility of estimating the hyper-parameters gives much more flexibility for semi-supervised methods. 
 
However, the full Bayesian computations can become very heavy computationally. In particular when the forward model is complex and the evaluation of the likelihood needs high computational cost. 
In those cases using surrogate simpler models can become very helpful to reduce the computational costs, but then, we have to account for uncertainty quantification (UQ) of the obtained results \cite{bayesian2018zhu}. Neural Networks (NN) with their diversity such as Convolutional (CNN), Deep learning (DL), etc., have become tools as fast and low computational surrogate forward models for them. 

In the last three decades, the Machine Learning (ML) methods and algorithms have gained great success in many computer vision (CV) tasks, such as classification, clustering, object detection, semantic segmentation, etc. These methods are mainly based on Neural Networks (NN) and in particular Convolutional NN (CNN), Deep NN, etc. \cite{review2017unser,9152164,9403414,9084378,9000801,8907811,using2018lucas,9403414,9084378,9000801}. 
 
Using these methods directly for inverse problems, as intermediate pre-processing or as tools for doing fast approximate computation in different steps of regularization or Bayesian inference have also got success, but not yet as much as they could. 
Recently, the Physics-Informed Neural Networks have gained great success in many inverse problems, proposing interaction between the Bayesian formulation of forward models,  optimization algorithms and ML specific algorithms for intermediate hidden variables. 
These methods have become very helpful to obtain approximate practical solutions to inverse problems in real world applications 
\cite{raissi2017physicsI,raissi2017physicsII,physicsinformed2019chen,physicsinformed2019raissi,9403414,8878159,9000801,8901171,8434321}.  

In this paper, first, in Section 2, a few general idea of ML, NN and DL are summarised, then in Section 3, 4 and 5, we focus on the NN and DL methods for inverse problems. First, we present same cases where we know the forward and its adjoint model. Then, we consider the case we may not have this knowledge and want to propose directly data driven DL methods    
\cite{1029279,AMD2021}.

\section{Machine Learning and Neural Networks basic idea}

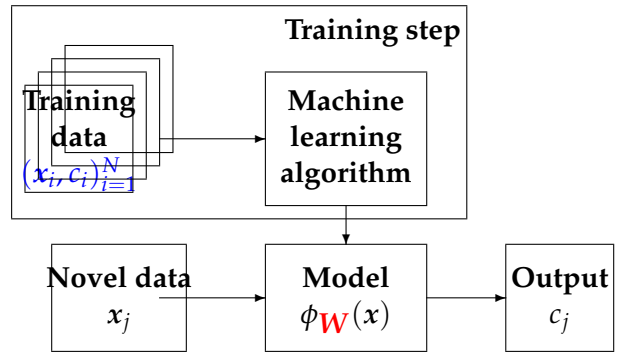
\begin{figure}[htb!]
\bcc
\begin{picture}(220,60)
  \put(0,0){\framebox(40,40){\bf\btabu{c} Training \\ data \\ $\blue{(\xb_i,c_i)_{i=1}^N}$ \etabu}}
  \put(5,5){\framebox(40,40){}}
  \put(10,10){\framebox(40,40){}}
  \put(15,15){\framebox(40,40){}}
  \put(-5,-10){\framebox(170,80){}}
  \put(120,50){\makebox(20,20){\bf Training step}}
  \put(90,-5){\framebox(60,50){\bf\btabu{c} Machine \\ learning \\ algorithm \etabu}}
  \put(50,20){\vector(1,0){40}}
  \put(120,-5){\vector(0,-1){15}}
  \put(10,-60){\framebox(50,40){\bf\btabu{c} Novel data \\ $\xb_j$ \etabu}}
  \put(50,-40){\vector(1,0){40}}
  \put(90,-60){\framebox(60,40){\bf\btabu{c} Model \\ $\phi_{\rWb}(\xb)$ \etabu}}
  \put(150,-40){\vector(1,0){30}}
  \put(180,-60){\framebox(40,40){\bf\btabu{c} Output \\ $c_j$ \etabu}}
\end{picture}
\vspace{1.7cm}
\ecc
\caption{Basic Machine Learning process: First Learn a model, then use it. Learning step needs a rich enough data base which costs a lot. When the model is learned and tested, its use is easy, fast and its cost is low.}
\label{fig01}
\end{figure}

The main idea in Machine Learning is first \emph{to learn} a model $f_{\rWb}(\xb)$ from a great number of input-output training data, for example, in a supervised classification problem, data classes: \qquad $(\xb_i,c_i), i=1,\cdots N$:

\vspace{-9pt}\[
\barr{c} \mbox{Learning Data}\\
\blue{(\xb_i,c_i)_{i=1}^N}
\earr
\ra
\fbox{\btabu{c} Learning step \\ The weights $\rWb$ of the NN model $\phi_{\rWb}(\xb)$ are obtained \etabu}
\ra
\rthetab, 
\]
and then, when a new case (Test $\xb_j$) appears, it uses the learned weights $\rWb$ to give a decision $c_j$. 

\[
\barr{c} \mbox{Test case Data}\\
\blue{\xb_j}
\earr
\ra
\fbox{\btabu{c} Testing step \\ The learned weights $\rWb$ are used in $\phi_{\rWb}(\xb)$\etabu}
\ra
\red{c_j}.
\]
Figure~\ref{fig01} shows the main process of ML. 

\medskip

\section{ML for inverse problems}
To show the possibilities of the interaction between inverse problems methods, Machine learning and NN methods, the best way is to give a few examples. 

\subsection{First example: A known linear forward model}
The first and easiest example is the case of linear inverse problems $\bgb=\Hb\rfb+\epsilonb$ where we know the forward model $\Hb$ and quadratic regularization where the solution is defined as: 
\beq
\rfbh=\argmin{\rfb}{\|\bgb-\Hb\rfb\|^2+\lambda\|\rfb\|^2},
\eeq
which has an analytic expression and we have the following relations:
\beq
\rfbh=(\Hb^t\Hb+\lambda \Ib)^{-1}\Hb^t \bgb = \Ab \, \bgb =\Bb \Hb^t \, \bgb
\mbox{~~or still~~} 
\rfbh=\Hb^t(\frac{1}{\lambda}\Hb\Hb^t+\Ib)^{-1}\bgb =\Hb^t \Cb \, \bgb,
\eeq
where $\Ab=(\Hb^t\Hb+\lambda \Ib)^{-1}\Hb^t$, $\Bb=(\Hb^t\Hb+\lambda \Ib)^{-1}$ and $\Cb=(\frac{1}{\lambda}\Hb\Hb^t+\Ib)^{-1}$. 

These relations can be presented schematically as: 
\[
\bgb\ra\fbox{$~~~~~~\Ab~~~~~~$}\ra\rfbh, \qquad 
\bgb\ra\fbox{$~~\Hb^t~~$}\ra\fbox{$~~\Bb~~$ }\ra\rfbh, \qquad 
\bgb\ra\fbox{$~~\Cb~~$}\ra\fbox{$~~\Hb^t~~$ }\ra\rfbh.
\]

As we can see, these relations directly induce linear feed forward  NN structure. In~particular, if~$\Hb$ represents a convolution operator, then $\Hb^t$, $\Hb^t\Hb$ and  $\Hb\Hb^t$ are too, as well as the operators $\Bb$ and $\Cb$.  Thus the whole inversion can be modelled by CNN~\cite{9152164,using2018lucas}. 

\begin{figure}[h!]
\bcc
\btabu{cc}
\btabu{c}
$\bgb\ra\fbox{$~~~~~~~~~\Ab~~~~~~~~~$}\ra\rfbh$ \\[6pt]
Totally data driven NN 
\etabu
&
\btabu{c}
\includegraphics[height=3cm,width=8cm]{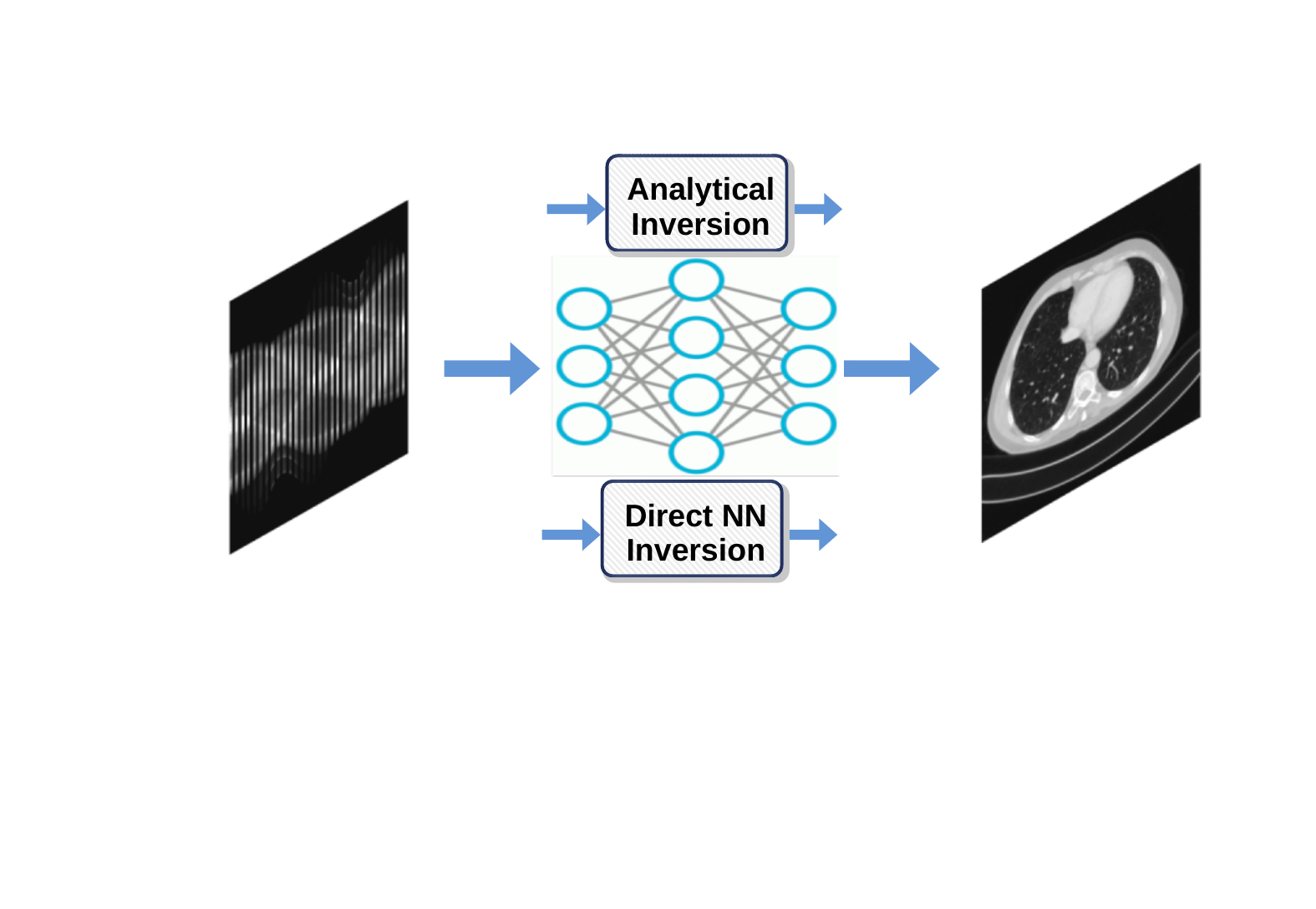}
\etabu
\\[-38pt]
\btabu{c}
$\bgb\ra\fbox{$~~\Hb^t~~$}\ra\fbox{$~~\Bb~~$ }\ra\rfbh$ \\[6pt]
Backprojection + NN 2D filtering
\etabu
&
\btabu{c}
\includegraphics[height=3cm,width=8cm]{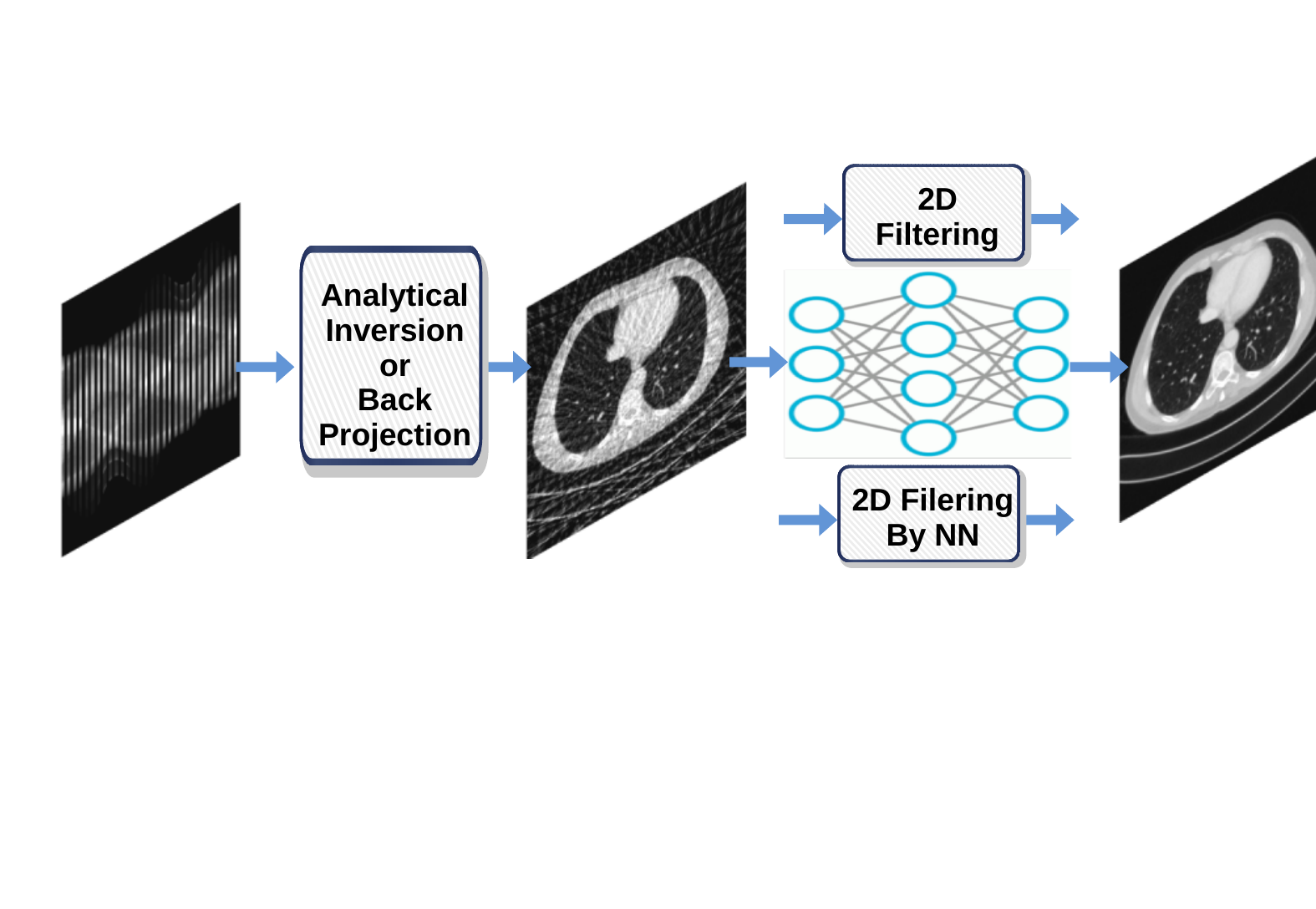} 
\etabu
\\[-38pt]
\btabu{c}
$\bgb\ra\fbox{$~~\Cb~~$}\ra\fbox{$~~\Hb^t~~$ }\ra\rfbh$ \\[6pt] 
NN 1D filtering + Back projection (FBP)
\etabu
&
\btabu{c}
\includegraphics[height=3cm,width=8cm]{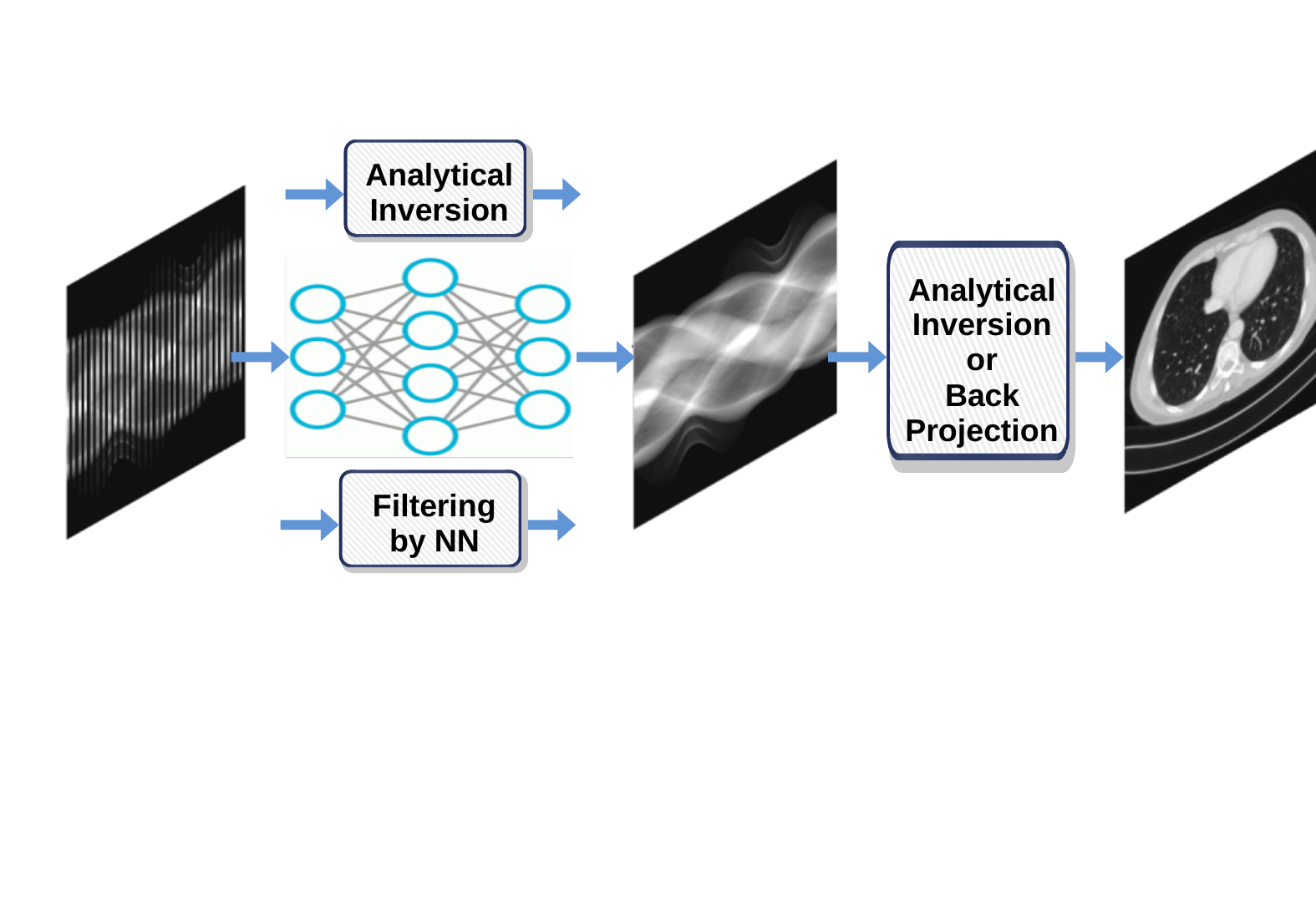}
\etabu
\\[-38pt]
\etabu
\ecc
\caption{Three linear NN structures which are derived directly from quadratic regularization inversion method. Right part of this figure is adapted from \cite{using2018lucas}.}
\label{fig02}
\end{figure}

For the case of Computed Tomography (CT), the first operation is equivalent to an analytic inversion, the second corresponds to Back-Projection first followed by 2D filtering in the image domain, and the third correspond to to the famous Filtered Back-Projection (FBP) which is implemented on classical CT scans.  
These three cases are illustrated on Figure~\ref{fig02}. 

\vspace{-18pt}

\subsection{Second example: Image denoising with a two layers CNN}
The second example is the denoising $\bgb=\rfb+\epsilonb$ with $\ell_1$ regularizer:
\beq
\rfbh=\Db\rzbh \mbox{~and~} \rzbh=\argmin{\rzb}{J(\rzb)} \mbox{~with~} 
J(\rzb)=\|\bgb-\Db\rzb|+\lambda\|\rzb\|_1 \, ,
\eeq
where $\Db$ is a filter, i.e., a convolution operator. 
This can also be considered as the~MAP estimator with a double exponential prior. 
It is easy to show that the solution can be obtained by a convolution followed by a thresholding~
\cite{learning2017meinhardt,regularized2019vettam}.
\beq
\rfbh=\Db\rzbh \mbox{~~and~~}
\rzbh=S_{\frac{1}{\lambda}}(\Db^t \bgb),
\eeq
where $S_\lambda$ is a thresholding operator. 
\[
\bgb\ra\fbox{$~~\Db^t~~$}\ra\fbox{Thresholding }\ra\rzbh\ra\fbox{$~~\Db~~$}\ra\rfbh
\mbox{~~or equivalently~~}
\bgb\ra\fbox{~~~Two layers CNN~~~ }\ra\rfbh.
\]

\subsection{Third example: A Deep learning equivalence of iterative gradient based algorithms}
One of the classical iterative methods in linear inverse problems algorithm is based on  the gradient descent method to optimize $J(\rfb)=\|\bgb-\Hb\rfb\|^2$: 
\beq
\rfbkp=\rfbk+\alpha\Hb^t(\bgb-\Hb\rfbk)=\alpha\Hb^t\bgb +(\Ib - \alpha\Hb^t\Hb)\rfbk\, ,
\eeq
where the solution of the problem is obtained recursively. Everybody knows that, when the forward model operator $\Hb$ is singular or ill-conditioned, this iterative algorithm starts by converging, but it may diverge easily. One of the experimental methods to obtain an acceptable approximate solution is just to stop the iterations after $K$ iterations. This idea can be translated to a Deep Learning NN by using $K$ layers. Each layer represents one iteration of the algorithm. See Figure~\ref{fig03}. 

\def\blocDL{\begin{picture}(100,80)
  \put(-10,20){\makebox(0,0){$\rfbk$}}
  \put(0,20){\vector(1,0){20}}

  \put(20,10){\framebox(50,20){$(\Ib - \alpha\Hb^t\Hb)$}}
  \put(70,20){\vector(1,0){20}}

  \put(100,80){\makebox(0,0){$\bgb$}}
  \put(100,80){\circle{20}}
  \put(100,70){\vector(0,-1){10}}
  \put(85,40){\framebox(30,20){$\alpha\Hb^t$}}  
  \put(100,40){\vector(0,-1){10}}

  \put(100,20){\circle{20}}
  \put(100,20){\makebox(0,0){$+$}}
  \put(110,20){\vector(1,0){20}}
  \put(135,20){\makebox(30,0){$\rfbkp$}}

  \put(10,0){\framebox(110,65){}}

\end{picture}
}

\def\blocDLa{\begin{picture}(100,80)
  \put(100,80){\makebox(0,0){$\bgb$}}
  \put(100,80){\circle{20}}
  \put(100,70){\vector(0,-1){10}}
  \put(85,40){\framebox(30,20){$\alpha\Hb^t$}}  
  \put(100,40){\vector(0,-1){10}}
\end{picture}
}

\def\blocDLah{\begin{picture}(100,80)
  \put(0,0){\makebox(20,20){$\bgb$}}
  \put(10,10){\circle{20}}
  \put(20,10){\vector(1,0){10}}
  \put(30,0){\framebox(30,20){$\alpha\Hb^t$}}  
  \put(60,10){\vector(1,0){10}}
\end{picture}
}

\def\blocDLb{\begin{picture}(100,80)
  \put(-10,20){\makebox(0,0){$\rfbk$}}
  \put(0,20){\vector(1,0){20}}

  \put(20,10){\framebox(50,20){$(\Ib - \alpha\Hb^t\Hb)$}}
  \put(70,20){\vector(1,0){20}}
  
  \put(100,20){\circle{20}}
  \put(100,20){\makebox(0,0){$+$}}
  \put(110,20){\vector(1,0){20}}
  \put(135,20){\makebox(30,0){$\rfbkp$}}
\end{picture}
}

\def\blocDLc{\begin{picture}(100,80)
  \put(0,0){\blocDLa}
  \put(0,0){\blocDLb}
  \put(10,0){\framebox(110,35){}}
\end{picture}
}

\def\blocDLd{\begin{picture}(0,0)
  \put(-5,20){\vector(1,0){15}}
  \put(10,10){\framebox(50,20){$(\Ib - \alpha\Hb^t\Hb)$}}
  \put(60,20){\vector(1,0){10}}
  \put(80,20){\circle{20}}
  \put(80,20){\makebox(0,0){$+$}}
  \put(80,45){\vector(0,-1){15}}
  \put(90,20){\vector(1,0){15}}
  \put(5,0){\framebox(90,35){}}
\end{picture}
}

\def\blocDLe{\begin{picture}(100,100)
  \put(-50,15){\blocDLa}
  \put(-100,0){\blocDLd}
  \put(5,0){\blocDLd}
  \put(115,20){\makebox(0,0){...}}
  \put(125,0){\blocDLd}
  \put(-105,45){\line(1,0){310}}
  \put(-105,45){\line(0,-1){25}}
  \put(-130,10){\makebox(20,20){$\rfb^{(1)}$}}
  \put(230,10){\makebox(20,20){$\rfb^{(K)}$}}
\end{picture}
}

\def\blocDLf{\begin{picture}(100,100)
  \put(-50,15){\blocDLah}
  \put(-100,0){\blocDLd}
  \put(5,0){\blocDLd}
  \put(115,20){\makebox(0,0){...}}
  \put(125,0){\blocDLd}
  \put(-105,45){\line(1,0){310}}
  \put(-105,45){\line(0,-1){25}}
\end{picture}
}

\begin{figure}[htb!]
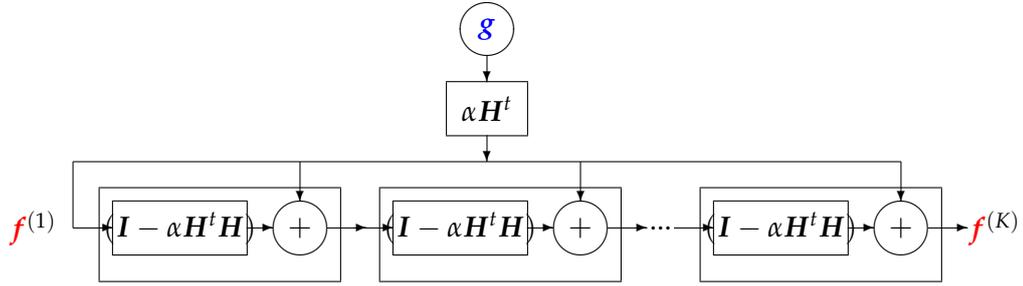

\bcc
\blocDLe
\ecc
\vspace{-9pt}
\caption{A $K$ layers DL NN equivalent to $K$ iterations of the basic optimization algorithm.}
\label{fig03}
\end{figure}
This DL structure can easily be extended to a regularized criterion: $J(\rfb)=\frac{1}{2}\|\bgb-\Hb\rfb\|^2+\lambda\|\Db\rfb\|^2$, where
\beq
\rfbkp=\rfbk+\alpha[\Hb^t(\bgb-\Hb\rfbk)-\lambda\Db^t\Db]=\alpha\Hb^t\bgb +(\Ib - \alpha\Hb^t\Hb-\alpha\lambda\Db^t\Db)\rfbk\, .
\eeq
We just need to replace $(\Ib - \alpha\Hb^t\Hb)$ by $(\Ib - \alpha\Hb^t\Hb-\alpha\lambda\Db^t\Db)$. 

This structure can also be extended to all the sparsity enforcing regularization terms such as $\ell_1$ and Total Variation (TV) using appropriate algorithms such as ISTA (Iterative Soft Thresholding Algorithm) or its fast version FISTA. by replacing the update expression and by adding a NL operation much like the ordinary NNs. 
A simple example is given in the following subsection. 

\subsection{Fourth example: $\ell_1$ regularization and NN}

Let us consider the linear inverse problem $\bgb=\Hb\rfb+\epsilonb$ with  $\ell_1$ regularization criterion:
\beq
J(\rfb) = \|\bgb-\Hb\rfb\|_2^2 + \lambda \|\rfb\|_1\, ,
\eeq
and an iterative optimization algorithm, such as ISTA: 
\beq
\hspace{-1cm}\rfb^{(k+1)} = 
Prox_{\ell_1} \left(\rfb^{(k)},\lambda\right) 
\defined 
\Sc_{\lambda\alpha}\left({\alpha}
\Hb^t\bgb + (\Ib - {\alpha} \Hb^t\Hb) \rfb^{(k)}\right)\, ,
\eeq
where $\Sc_{\theta}$ is a soft thresholding operator and  
$\alpha \le |eig(\Hb^t\Hb)|$ is the Lipschitz constant of the normal operator. When $\Hb$ is a convolution operator, then:
\bit
\item $(\Ib - {\alpha} \Hb^t\Hb) \rfb^{(k)}$ 
can also be approximated by a convolution and thus considered as a filtering operator;
\item $\frac{1}{\alpha}\Hb^t\bgb$ can be considered as a bias term and is also a convolution operator; and 
\item $\Sc_{\theta=\lambda\alpha}$ is a nonlinear point wise operator. In particular when $\rfb$ is a positive quantity, this soft thresholding operator can be compared to ReLU activation function of NN. See Figure~\ref{fig04}.
\eit

\def\SoftT{
\bpic(25,20)
  \put(0,10){\line(1,0){20}}
  \put(10,0){\line(0,1){20}}
  \put(15,10){\line(1,1){10}}
  \put(-5,0){\line(1,1){10}}
\epic
}

\def\Blocka{
\bpic(150,70)
  \put(0,20){\makebox(20,0){$\rfb^{(k)}$}}
  \put(20,20){\vector(1,0){10}}
  \put(30,10){\framebox(60,20){$(\Ib - {\alpha} \Hb^t\Hb)$}}
  \put(90,20){\vector(1,0){10}}
  \put(110,20){\circle{20}}
  \put(110,20){\makebox(0,0){$+$}}
  \put(120,20){\vector(1,0){10}}
  \put(130,10){\framebox(30,20){\SoftT}}
  \put(160,20){\vector(1,0){10}}
  \put(175,20){\makebox(20,0){$\rfb^{(k+1)}$}}
  \put(110,40){\vector(0,-1){10}}
  \put(100,40){\framebox(20,15){${\alpha}\Hb^t$}}
  \put(110,67){\vector(0,-1){10}}
  \put(105,70){\makebox(10,10){$\bgb$}}
  \put(110,75){\circle{20}}
\epic
}

\begin{figure}[htb!]
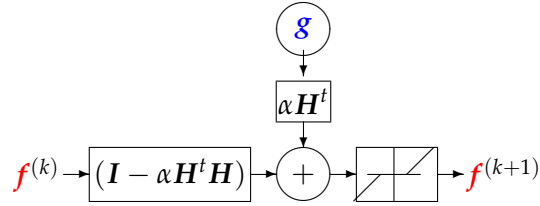

\bcc
\Blocka
\ecc
\vspace{-18pt}
\caption{One block of a NN correspond to one iteration of $\ell_1$ regularization.}
\label{fig04}
\end{figure}

\subsubsection{DL structure based on iterative inversion algorithm}
Using the iterative gradient based algorithm with fixed number of iterations for computing a GI or a regularized one as explained in previous section can be used to propose a DL structure with $K$ layers, $K$ being the number of iterations before stopping. Figure~\ref{fig05} shows this structure for a quadratic regularization which results to a linear NN and Figure~\ref{fig06} for the case of $\ell_1$  regularization.  

\def\blocDLf{\begin{picture}(100,50)
  \put(-175,10){\blocDLah}
  \put(-100,0){\blocDLd}
  \put(5,0){\blocDLd}
  \put(115,20){\makebox(0,0){...}}
  \put(125,0){\blocDLd}
  \put(-105,45){\line(1,0){310}}
  \put(-105,45){\line(0,-1){25}}
  \put(240,20){\makebox(0,0){$\rfbh$}}
\end{picture}
}

\def\Blockw{
\bpic(150,50)
  \put(0,20){\makebox(20,0){$\rfb^{(k)}$}}
  \put(20,20){\vector(1,0){10}}
  \put(30,10){\framebox(60,20){$\rWb^{(k)}$}}
  \put(90,20){\vector(1,0){10}}
  \put(110,20){\circle{20}}
  \put(110,20){\makebox(0,0){$+$}}
  \put(120,20){\vector(1,0){10}}
  \put(130,10){\framebox(30,20){\SoftT}}
  \put(160,20){\vector(1,0){10}}
  \put(175,20){\makebox(20,0){$\rfb^{(k+1)}$}}
  \put(110,40){\vector(0,-1){10}}
  \put(100,40){\framebox(20,15){${\rWb_0}$}}
  \put(110,67){\vector(0,-1){10}}
  \put(105,70){\makebox(10,10){$\bgb$}}
  \put(110,75){\circle{20}}
\epic
}

\medskip
\begin{figure}[htb!]
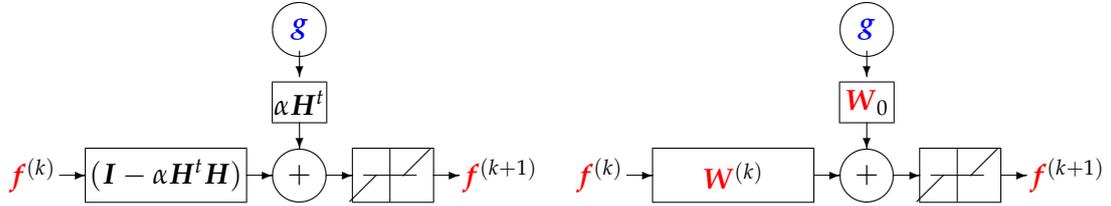

\bcc
\hspace*{-15mm}\Blocka \hspace{2cm} \Blockw
\ecc
\caption{A $K$ layers DL NN equivalent to $K$ iterations of a basic gradient based optimization algorithm. A quadratic regularization results to a linear NN while a $\ell_1$ regularization results to a classical NN with a nonlinear activation function. Left: supervised case. Right: unsupervised case. In both cases, all the $K$ layers have the same structure.}
\label{fig05}
\end{figure}

\def\Blockwx#1#2{
\bpic(150,80)
  \put(20,20){\vector(1,0){10}}
  \put(30,10){\framebox(30,20){#2}}
  \put(60,20){\vector(1,0){10}}
  \put(80,20){\circle{20}}
  \put(80,20){\makebox(0,0){$+$}}
  \put(90,20){\vector(1,0){10}}
  \put(100,10){\framebox(30,20){\SoftT}}
  \put(130,20){\vector(1,0){10}}
  \put(80,40){\vector(0,-1){10}}
  \put(70,40){\framebox(20,15){#1}}
  \put(80,67){\vector(0,-1){10}}
  \put(75,70){\makebox(10,10){$\bgb$}}
  \put(80,75){\circle{20}}
\epic
}
\def\BlockDLw{\begin{picture}(100,100)
  \put(-160,10){\Blockwx{$\rWb_0$}{$\rWb^{(1)}$}}
  \put(-40,10){\Blockwx{$\rWb_0$}{$\rWb^{(2)}$}}
  \put(90,10){\Blockwx{$\rWb_0$}{$\rWb^{(K)}$}}
  \put(108,30){\makebox(0,0){...}}
  \put(-140,35){\makebox(0,0){$\rfbh^{(1)}$}}
  \put(245,35){\makebox(0,0){$\rfbh^{(K)}$}}
\end{picture}
}

\begin{figure}[htb!]
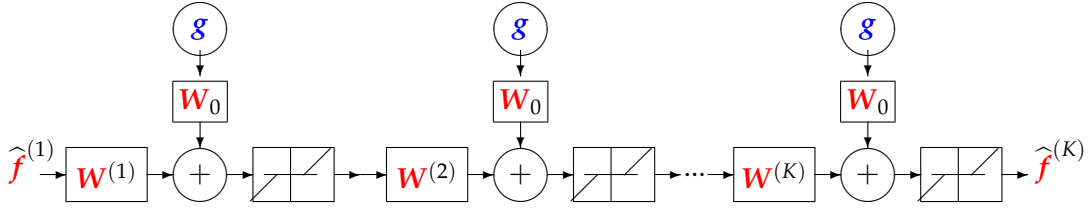

\bcc
\BlockDLw
\ecc
\vspace{-18pt}
\caption{All the $K$ layers of DL NN equivalent to $K$ iterations of an iterative gradient based optimization algorithm. The simplest solution is to choose 
$\rWb_0=\alpha\Hb$ ~~ and  ~~
$\rWb^{(k)}=\rWb=(\Ib-\alpha\Hb^t\Hb), \; k=1,\cdots,K$.  
A more robust, but more costly is to learn all the layers $\rWb^{(k)}=(\Ib-\alpha^{(k)}\Hb^t\Hb), \quad k=1,\cdots,K$.}
\label{fig06}
\end{figure}

In all these examples, we directly could obtain the structure of the NN from the Forward model and known parameters. However, in these approaches there are some difficulties which consist in the determination of the structure of the NN. For example, in the first example, obtaining the structure of $\Bb$ depends on the regularization parameter $\lambda$. The same difficulty arises for determining the shape and the threshold level of the Thresholding bloc of the network in the second example. 
The same need of the regularization parameter as well as many other hyper parameters are necessary to create the NN structure and weights. 
In practice, we can decide, for example, on the number and structure of a DL network, but as their corresponding weights depend on many unknown or difficult to fix parameters, ML may become of help.  In the following we first consider 
the training part of a general ML method. Then, we will see how to include the physics based knowledge of the forward model in the structure of learning.

\section{More Physics based ML using linear transformations}
As mentioned above, in general, in practice, a rich enough and complete data set is not often available in particular for inverse problems. We have, as far as possible, to use the physics of the forward operator $\Hc$. Sometimes, the forward operator can be described in a transform domain, such Fourier or Wavelets. Here, we explore these situations. 

\subsection{Decomposition of the NN structure to fixed and trainable parts}
\begin{figure}[htb!]
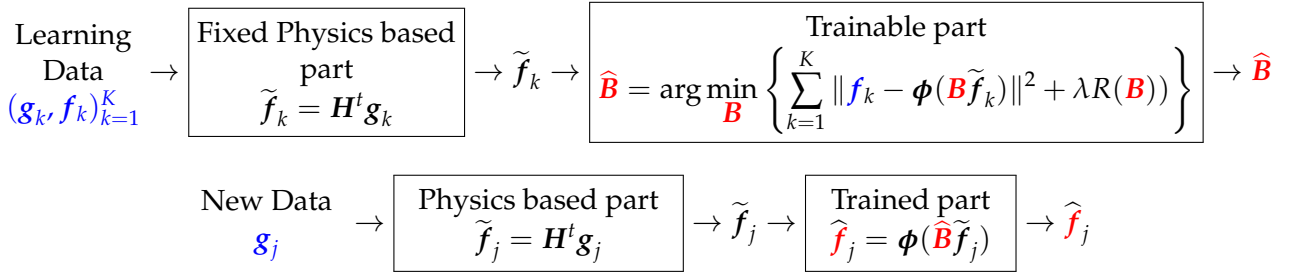

\[
\barr{@{}c@{}} 
\mbox{Learning} \\ \mbox{Data}\\
\blue{(\gb_k,\bfb_k)_{k=1}^K}
\earr
\ra
\fbox{\btabu{@{}c@{}} Fixed Physics based \\ part \\ $\fbt_k=\Hb^t\gb_k$  \etabu}
\ra
\fbt_k
\ra
\fbox{\btabu{@{}c@{}} Trainable part \\ 
$\disp{\rBbh=\argmin{\rBb}{\sum_{k=1}^K \|\bfb_k-\phib(\rBb\fbt_k)\|^2+\lambda R(\rBb))}}$  \etabu}
\ra
\rBbh
\]
\[
\barr{c} \mbox{New Data}\\
\blue{\gb_j}
\earr
\ra
\fbox{\btabu{c} Physics based part \\ $\fbt_j=\Hb^t\gb_j$  \etabu}
\ra
\fbt_j
\ra
\fbox{\btabu{c} Trained part \\ $\rfbh_j=\phib(\rBbh \fbt_j)$ \etabu}
\ra
\rfbh_j
\]
\caption{Training (top) and Testing (bottom) steps in the first use of physics based ML approach}
\label{fig07}
\end{figure}

The first easiest and understandable method consists in decomposing the structure of the network $\rWb$ in two parts: a fixed part and a learnable part. As the simplest example, we can consider the case of analytical expression of the quadratic regularization: 
$\rfbh=(\Hb\Hb^t+\lambda \Db\Db^t)^{-1}\Hb^t \bgb=\Bb \Hb^t \bgb$ which suggests to have a two layers network with a fixed part structure $\Hb^t$ and a trainable one $\Bb=(\Hb\Hb^t+\lambda \Db\Db^t)^{-1}$. 
See Figure~\ref{fig07}. 

It is interesting to note that in X-ray Computed Tomography (CT) the forward operator $\Hb$ is called \emph{Projection}, the adjoint operator $\Hb^t$ is called \emph{Back-Projection (BP)} and the $\Bb$ operator is assimilated to a 2D filtering (convolution). 

\subsection{Using Singular value decomposition of forward and backward operators}

\begin{figure}[htb!]
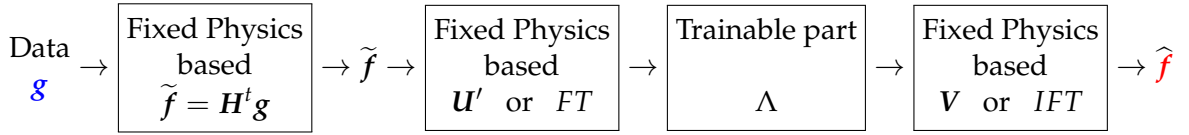

\[
\barr{@{}c@{}} \mbox{Data}\\
\bgb
\earr
\ra
\fbox{\btabu{@{}c@{}} Fixed Physics \\ based \\ $\fbt=\Hb^t\gb$  \etabu}
\ra
\fbt
\ra
\fbox{\btabu{@{}c@{}} Fixed Physics \\ based \\ $\Ub'$ ~~or~~ $FT$  \etabu}
\ra\fbox{\btabu{@{}c@{}} Trainable part \\ ~\\ $\Lambda$  \etabu}
\ra
\fbox{\btabu{@{}c@{}} Fixed Physics \\ based \\ $\Vb$ ~~or~~ $IFT$  \etabu}
\ra
\rfbh
\]
\caption{A four-layers NN with three physics based fixed corresponding to $\Hb^t$, $\Ub'$ or $FT$ and $\Vb$ or $IFT$ layers and one trainable layer corresponding to $\Lambda$.}
\label{fig08}
\end{figure}

Using the eigenvalues and eigenvectors of the pseudo or generalized inverse operators 
\beq
\Hb^\dag=[\Hb^t\Hb]^{-1}\Hb^t 
\quad\mbox{~~~or~~~}\quad 
\Hb^\dag=\Hb^t[\Hb\Hb^t]^{-1},
\eeq
and Singular value decomposition (SVD) of the operators $[\Hb^t\Hb]$ and $[\Hb\Hb^t]$ give another possible decomposition of the NN structure. Let us to note 
\beq
\Hb\Hb^t=\Ub \Deltab \Vb' \mbox{~or equivalently~} \Hb^t\Hb=\Vb \Deltab \Ub'\, ,
\eeq
where $\Deltab$ is a diagonal matrix containing the singular values, $\Ub$ and $\Vb$ containing the corresponding eigenvectors. This can be used to decompose the $\rWb$ to four operators:
\beq
\rWb = \Vb'\Deltab \Ub \Hb^t 
\quad\mbox{~or~}\quad 
\rWb = \Hb^t \Vb \Deltab \Ub'\, ,
\eeq
where three of them can be fixed and only one $\Deltab$ can be trainable. It is interesting to know that when the forward operator $\Hb$ has a shift-invariant (convolution) property, then the operators $\Ub$ and $\Vb'$ will correspond, respectively, to the FT and IFT operators and the diagonal elements of $\Lambda$ correspond to the FT of the impulse response of the convolution forward operator. 
So, we will have a fixed layer corresponding to $\Hb^t$ which can be interpreted as a matched filtering, then a fixed FT layer which is a feed-forward linear network, a trainable filtering part corresponding to the diagonal elements of $\Lambdab$ and a forth fixed layer corresponding to IFT. 
See Figure~\ref{fig08}. 

\section{Learning step general approach}
The ML approach can become helpful if we could have a great amount of data: 
inputs-outputs \\ 
$\left\{(\bfb,\bgb)_k, k=1,2, ..., K\right\}$ examples. Thus, during the Training step, we can learn the coefficients of the NN and then use it for obtaining a new solution $\rfbh$ for a new data $\bgb$.

The main issue is the limited number of data input-output examples  $\left\{(\bfb,\bgb)_k, k=1,2, ..., K\right\}$ we can have for the training step of the network. 

\subsection{Fully learned method}
Let consider a one layer NN where the relation between its input $\bgb_k$ and output $\bfb_k$ is given by $\bfb_k=\phib(\rWb\bgb_k)$ where $\rWb$ is the weighting parameters of the NN and $\phib$ is the point wise  non linearity function of the output NN output layer. The  estimation of $\rWb$ from the training data in the learning step is done by an optimization algorithm which optimizes a Loss function $\Lc$ defined as
\beq
\Lc(\rWb) = \sum_{k=1}^K \ell_k(\bfb_k,\phib(\rWb\bgb_k))+\lambda \|\rWb\|^2
\mbox{~~with~~}
\ell_k(\bfb_k,\phib(\rWb\bgb_k)=\|\bfb_k-\phib(\rWb\bgb_k)\|^2\, ,
\eeq
a quadratic distance or any other appropriate distance or divergence or a probabilistic one \\ 
\(
\ell_k(\bfb_k,\phib(\rWb\bgb_k)=\esp{\|\bfb_k-\phib(\rWb\bgb_k)\|^2}\, ,
\)
and where $\|\rWb\|^2$ is a regularizing term and $\lambda$ its parameter. 

\begin{figure}[htb!]
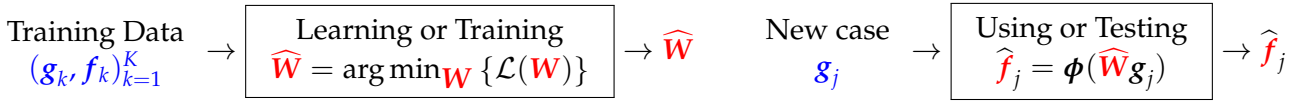

\[
\barr{c} \mbox{Training Data}\\
\blue{(\gb_k,\bfb_k)_{k=1}^K}
\earr
\ra
\fbox{\btabu{c} Learning or Training \\ $\rWbh=\argmin{\rWb}{\Lc(\rWb)}$  \etabu}
\ra
\rWbh
\qquad 
\barr{c} \mbox{New case}\\
\blue{\gb_j}
\earr
\ra
\fbox{\btabu{c} Using or Testing \\ $\rfbh_j=\phib(\rWbh \gb_j)$ \etabu}
\ra
\rfbh_j
\]
\caption{Training (top) and Testing (bottom) steps in a ML approach}
\label{fig09}
\end{figure}

When the NN is trained and we obtain the weights $\rWbh$, then we can use it easily when a new case (Test $\bgb_j$) appears, just by applying: $\rfb_k=\phib(\rWbh\bgb_k)$. 
These two steps of Training and Using (called also Testing) are illustrated in Figure~\ref{fig09}. 

The scheme that we presented is general and can be extended to any multi-layer NN and DL. 
In fact, if we had a great number of data-ground truth examples $\left\{(\rfb,\bgb)_k, k=1,2, ..., K\right\}$ with $K$ much more than the number of elements $W_{m,n}$ of the weighting parameters $\rWb$, then, we did not even have any need for forward model $\Hc$. This can be possible for very low dimensional problems \cite{using2018lucas}. 
But, in general, in practice we do not have enough data. So, some prior or regularizer is needed to obtain a usable solution. 

\section{Application: Infrared imaging}
In many industrial application, Infrared (IR) imaging is used to diagnosis and to survey the temperature field distribution of the objects. Two great difficulties with these images are: low resolution and important noise. To increase the resolution, we may use deconvolution methods if we can get the point spread function (PSF) of the camera. A solution to reduce the noise can also be obtained via a total variation prior modeling. Indeed, both objectives can be reached via a regularization or the Bayesian approach. Also, as the final objective is to segment image to obtain different levels of temperature (background, normal, high, and very high), we propose to design a BDL NN which gets as input a low resolution and noisy image and outputs a segmented image 
with 3 or 4 levels. 

To train this NN, we can generate different known shaped synthetic images to consider as the ground truth and simulate the blurring effects of temperature diffusion, via the convolution of different appropriate point spread functions and add some noise to generate realistic images. We can also use a black body thermal sources, for which we know the shape and the exact temperature, and acquire different images at different conditions. All these images can be used for the training of the network. 

We propose then to use a four groups of layers DL structure as it is shown in Figure~\ref{fig10}, to train it with one hundred images artificially generated and one hundred images obtained with a black body experiment. Then, trained model can be used for the desired task on a test set images. Figure~\ref{fig11}, we show one such expected results. More details will be given in a near future paper. 

\begin{figure}
\[
\barr{@{}c@{}c@{}c@{}l@{}c@{}}
\mbox{Input~} \bgb & denoising & deconv & Segmentation& \mbox{Segmented}\\
\mbox{IR image}&\ra\fbox{C1}-\fbox{Th}-\fbox{C2}\ra &
\fbox{C3}-\fbox{Thr}-\fbox{C4}\ra
&\fbox{~~~~SegNet~~~~}\ra &
\mbox{image}
\earr
\]
\caption{The proposed 4 layers NN for denoising, deconvolution and segmentation of IR images}
\label{fig10}
\end{figure}

\begin{figure}
\bcc
\btabu{ccc}
\includegraphics[height=4.5cm]{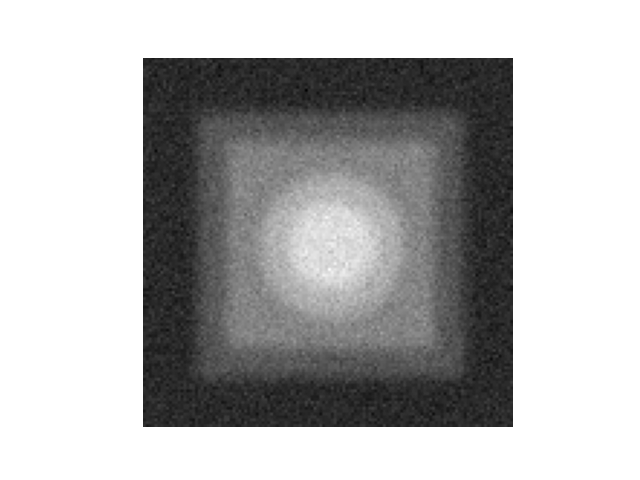}& \hspace{-1cm}
\includegraphics[height=4.5cm]{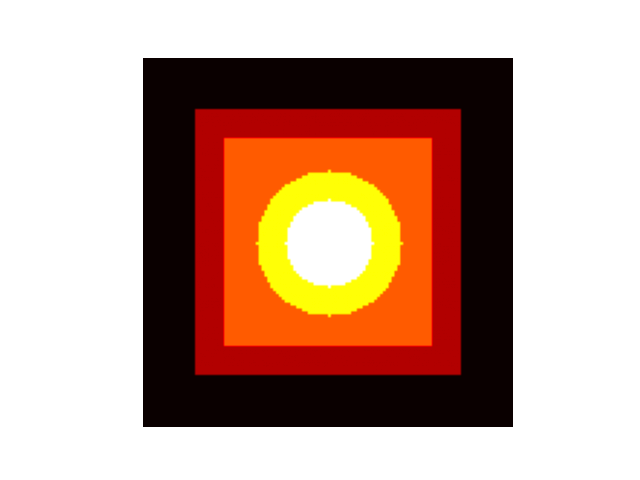}& \hspace{-1cm}
\includegraphics[height=4.5cm]{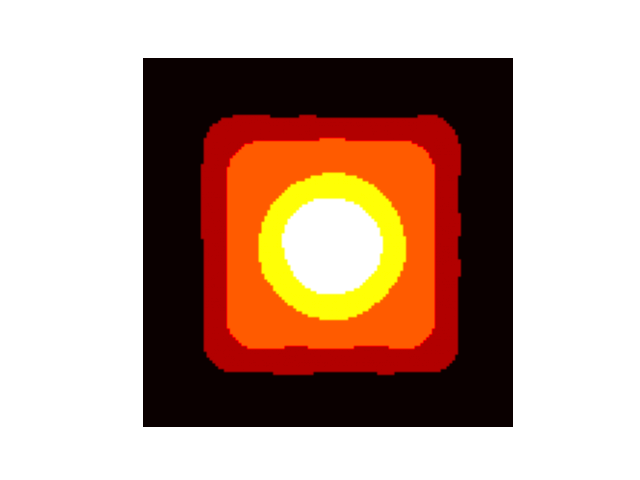}
\\ 
\includegraphics[height=4.5cm,width=6cm]{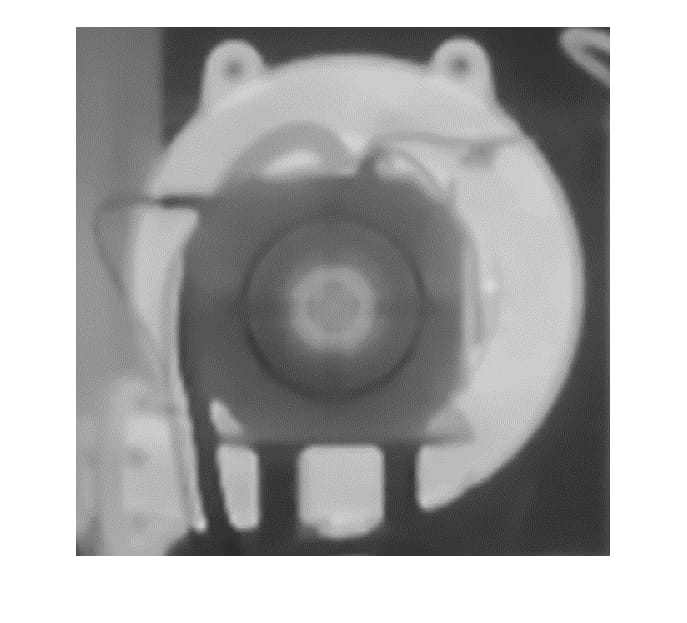}& & \hspace{-1cm}

\includegraphics[height=4.5cm,width=5cm]{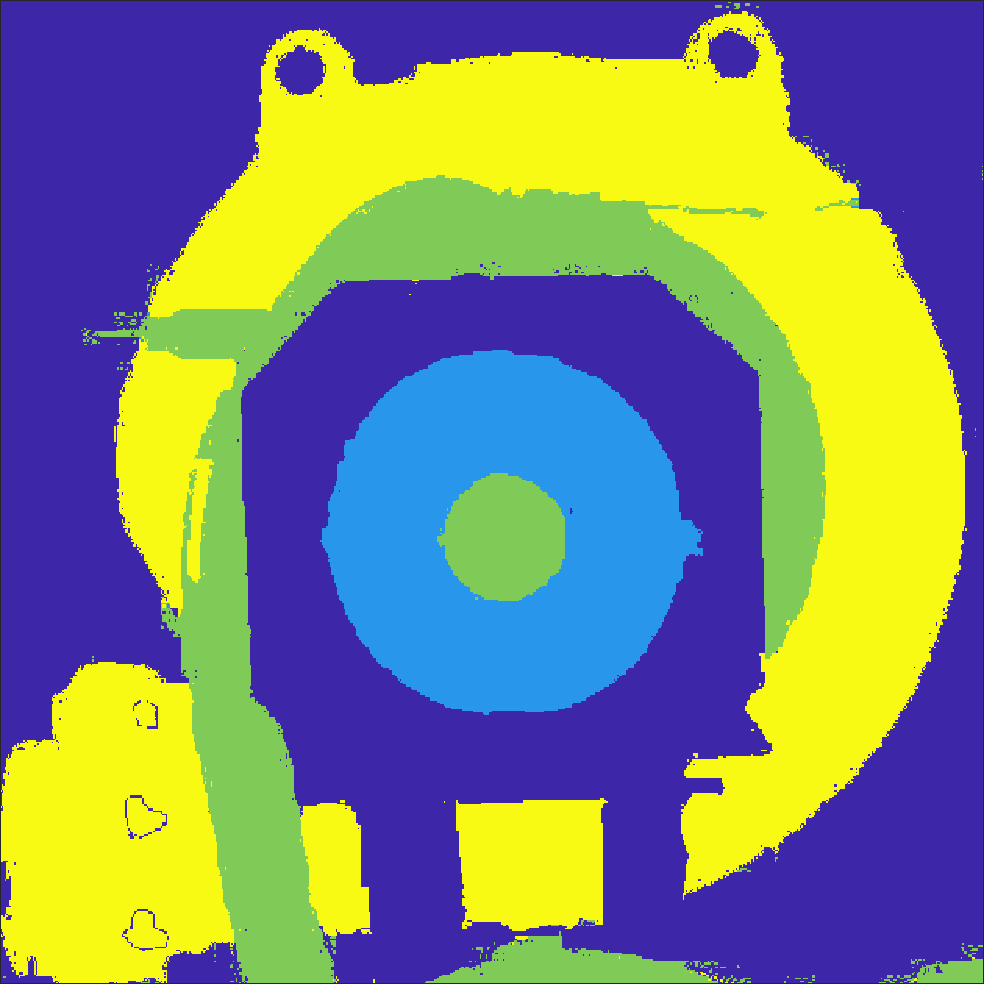} \\ 
\etabu
\ecc
\vspace{-12pt}
\caption{Example of expected results: First row: a simulated IR image (left), its ground truth labels (middle) , the result of the deconvolution and segmentation (right). Second row: a real IR image (left) and the result of its deconvolution and segmentation (right).}
\label{fig11}
\end{figure}

\section{Conclusions and~Challenges}
Classical methods for inverse problems are mainly based on regularization methods or on Bayesian inference with a connection between them via the Maximum A Posteriori (MAP) point estimation. The Bayesian approach gives more flexibility, in particular for determination of the regularization parameter. However, whatever deterministic or Bayesian computations still is a great problem for high dimensional problems. 

Recently, the Machine Learning (ML) methods have become a good help for some aspects of these difficulties. Nowadays, 
ML, Neural Networks (NN), Convolutional NN (CNN) and Deep Learning (DL) methods have obtained great success in classification, clustering, object detection, speech and face recognition, etc., But, they need a great number of training data and and they may fail very easily, in particular for inverse problems. 

In fact, using only data based NN without any specific structure coming from the forward model (Physics) may work for small size problems. However, the progress arrives via their interaction with the model based methods. In fact, the success of CNN and DL methods greatly depends on the appropriate choice of the network structure. This choice can be guided by the model based methods \cite{using2018lucas,learning2017meinhardt,survey2018guidotti,
regularized2019vettam,review2017jin,review2017unser,bayesian2018zhu,one2017chang,deep2018mo}. 

In this work, we presented a few examples of such interactions. We explored a few cases: first when the forward operator is known. Then, when we use the forward model partially or in the transform domain. As we could see, the main contribution of ML and NN tools can be in reducing the costs of the inversion method when an appropriate model is trained. However, to obtain a good model, there is a need for sufficiently rich data and a good network structure obtained from the physics knowledge of the problem in hand. 

For inverse problems, when the forward models are non linear and complex, NN and DL may be of great help. However, we may still need to choose the structure of the NN via an approximate forward model and approximate Bayesian inversion \cite{raissi2017physicsI,raissi2017physicsII,physicsinformed2019raissi}. 

\def\url#1{\tt#1}

\bibliographystyle{Inputs/ieeetr.bst}
\bibliography{biblio/IEEEXplore2020,biblio/IP_ML_IEEE,biblio/amd_2022,biblio/litmaps_ML_IP,biblio/references}

\end{document}

%% file: macros_gpi.tex
\def\bdoc{\begin{document}}             \def\edoc{\end{document}}
\def\babs{\begin{abstract}}             \def\eabs{\end{abstract}}
\def\bcc{\begin{center}}                \def\ecc{\end{center}}
\def\ben{\begin{enumerate}}             \def\een{\end{enumerate}}
\def\bit{\begin{itemize}}               \def\eit{\end{itemize}}
\def\bpic{\begin{picture}}              \def\epic{\end{picture}}
\def\barr{\begin{array}}                \def\earr{\end{array}}
\def\btabu{\begin{tabular}}             \def\etabu{\end{tabular}}
\def\bcc{\begin{center}}                \def\ecc{\end{center}}
\def\beq{\begin{equation}}              \def\eeq{\end{equation}}
\def\beqn{\begin{eqnarray}}             \def\eeqn{\end{eqnarray}}
\def\beqnx{\begin{eqnarray*}}           \def\eeqnx{\end{eqnarray*}}
\def\beqnn{\begin{eqnarray*}}			\def\eeqnn{\end{eqnarray*}}
\def\bfig{\begin{figure}}               \def\efig{\end{figure}}
\def\bver{\begin{verb}}					\def\ever{\end{verb}}
\def\bbib{\begin{thebibliography}{999}}	\def\ebib{\end{thebibliography}}

\def\blue#1{{\color{blue}#1}}
\def\red#1{{\color{red}#1}}
\def\green#1{{\color{green}#1}}

\def\wt{\widetilde}
\def\wh{\widehat}

\def\zerob{{\bm 0}}
\def\oneb{{\bm 1}}

\def\ab{{\bm a}}
\def\bb{{\bm b}}
\def\cb{{\bm c}}
\def\db{{\bm d}}
\def\eb{{\bm e}}
\def\fb{{\bm f}}
\def\gb{{\bm g}}
\def\hb{{\bm h}}
\def\ib{{\bm i}}
\def\jb{{\bm j}}
\def\kb{{\bm k}}
\def\lb{{\bm l}}
\def\mb{{\bm m}}
\def\nb{{\bm n}}
\def\ob{{\bm o}}
\def\pb{{\bm p}}
\def\qb{{\bm q}}
\def\rb{{\bm r}}
\def\sb{{\bm s}}
\def\tb{{\bm t}}
\def\ub{{\bm u}}
\def\vb{{\bm v}}
\def\wb{{\bm w}}
\def\xb{{\bm x}}
\def\yb{{\bm y}}
\def\zb{{\bm z}}

\def\Ab{{\bm A}}
\def\Bb{{\bm B}}
\def\Cb{{\bm C}}
\def\Db{{\bm D}}
\def\Eb{{\bm E}}
\def\Fb{{\bm F}}
\def\Gb{{\bm G}}
\def\Hb{{\bm H}}
\def\Ib{{\bm I}}
\def\Jb{{\bm J}}
\def\Kb{{\bm K}}
\def\Lb{{\bm L}}
\def\Mb{{\bm M}}
\def\Nb{{\bm N}}
\def\Ob{{\bm O}}
\def\Pb{{\bm P}}
\def\Qb{{\bm Q}}
\def\Rb{{\bm R}}
\def\Sb{{\bm S}}
\def\Tb{{\bm T}}
\def\Ub{{\bm U}}
\def\Vb{{\bm V}}
\def\Wb{{\bm W}}
\def\Xb{{\bm X}}
\def\Yb{{\bm Y}}
\def\Zb{{\bm Z}}

\def\alphab{\bm{\alpha}}
\def\betab{\bm{\beta}}
\def\deltab{\bm{\delta}}
\def\etab{\bm{\eta}}
\def\epsilonb{\bm{\epsilon}}
\def\gammab{\bm{\gamma}}
\def\omegab{\bm{\omega}}
\def\etab{\bm{\eta}}
\def\thetab{\bm{\theta}}
\def\xib{\bm{\xi}}
\def\lambdab{\bm{\lambda}}
\def\taub{\bm{\tau}}
\def\phib{\bm{\phi}}
\def\mub{\bm{\mu}}
\def\psib{\bm{\psi}}
\def\chib{\bm{\chi}}
\def\sigmab{\bm{\sigma}}

\def\Deltab{\bm{\Delta}}
\def\Lambdab{\bm{\Lambda}}
\def\Phib{\bm{\Phi}}
\def\Psib{\bm{\Psi}}
\def\Sigmab{\bm{\Sigma}}

\def\Ac{{\cal A}}
\def\Bc{{\cal B}}
\def\Cc{{\cal C}}
\def\Dc{{\cal D}}
\def\Ec{{\cal E}}
\def\Fc{{\cal F}}
\def\Gc{{\cal G}}
\def\Hc{{\cal H}}
\def\Ic{{\cal I}}
\def\Jc{{\cal J}}
\def\Kc{{\cal K}}
\def\Lc{{\cal L}}
\def\Mc{{\cal M}}
\def\Nc{{\cal N}}
\def\Oc{{\cal O}}
\def\Pc{{\cal P}}
\def\Qc{{\cal Q}}
\def\Rc{{\cal R}}
\def\Sc{{\cal S}}
\def\Tc{{\cal T}}
\def\Uc{{\cal U}}
\def\Vc{{\cal V}}
\def\Wc{{\cal W}}
\def\Xc{{\cal X}}
\def\Yc{{\cal Y}}
\def\Zc{{\cal Z}}

\def\d#1{\,\mbox{d} #1}
\def\disp#1{\displaystyle{#1}}

\def\iii{\int_{-\infty}^{+\infty}}
\def\izi{\int_{0}^{\infty}}
\def\izpi{\int_{0}^{\pi}}
\def\izdpi{\int_{0}^{2\pi}}
\def\intd{\int\kern-.8em\int}
\def\intt{\int\kern-.8em\int\kern-.8em\int}
\def\intg{\int\kern-1.1em\int}

\def\xh{\widehat{x}}
\def\thetah{\widehat{\theta}}
\def\betah{\widehat{\beta}}

\def\xbh{\widehat{\xb}}
\def\ybh{\widehat{\yb}}
\def\thetabh{\widehat{\thetab}}
\def\etabh{\widehat{\etab}}
\def\betabh{\widehat{\betab}}

\def\xbhk{\widehat{\xb}^{k}}
\def\thetahk{\widehat{\theta}^{k}}
\def\betahk{\widehat{\beta}^{k}}

\def\xbhkp{\widehat{\xb}^{k+1}}
\def\thetahkp{\widehat{\theta}^{k+1}}
\def\betahkp{\widehat{\beta}^{k+1}}

\def\thetabhk{\widehat{\thetab}^{k}}
\def\betabhk{\widehat{\betab}^{k}}

\def\thetabhkp{\widehat{\thetab}^{k+1}}
\def\betabhkp{\widehat{\betab}^{k+1}}

\def\thetamin{\theta_{\mbox{\tiny min}}}
\def\thetamax{\theta_{\mbox{\tiny max}}}
\def\betamin{\beta_{\mbox{\tiny min}}}
\def\betamax{\beta_{\mbox{\tiny max}}}

\def\ra{\rightarrow}
\def\la{\leftarrow}
\def\da{\downarrow}
\def\ua{\uparrow}

\def\Ra{\Rightarrow}
\def\La{\Leftarrow}
\def\Da{\Downarrow}
\def\Ua{\Uparrow}

\def\lra{\longrightarrow}
\def\lla{\longleftarrow}
\def\Lra{\Longrightarrow}
\def\Lla{\longleftarrow}

\def\lrarr{\leftrightarrow}
\def\Lrarr{\Leftrightarrow}
\def\udarr{\updownarrow}
\def\Uparr{\Updownarrow}

\def\expf#1{\exp\left[ {#1} \right]}

\def\argmax#1#2{\arg\max_{#1}\left\{ #2 \right\}}
\def\argmin#1#2{\arg\min_{#1}\left\{ #2 \right\}}

\def\Prob#1{\mbox{Pr}\left\{#1\right\}}
\def\var#1{\mbox{Var}\left\{#1\right\}}
\def\cov#1{\mbox{Cov}\left\{#1\right\}}
\def\corr#1{\mbox{Corr}\left\{#1\right\}}
\def\trace#1{\mbox{Tr}\left\{#1\right\}}
\def\rang#1{\mbox{rang}\left\{#1\right\}}
\def\det#1{\mbox{d\'et}\left\{#1\right\}}

\def\gbu{\underline{\gb}}
\def\fbu{\underline{\fb}}
\def\zbu{\underline{\zb}}
\def\epsilonbu{\underline{\epsilonb}}
\def\thetabu{\underline{\thetab}}

\def\ve{{v_{\epsilon}}}
\def\sigmae{{\sigma_{\epsilon}}}
\def\Sigmae{{\Sigmab_{\epsilonb}}}
\def\Sigmaf{{\Sigmab_{\fb}}}
\def\Sigmag{{\Sigmab_{\gb}}}

\def\fbuh{\widehat{\fbu}}
\def\zbuh{\widehat{\fbu}}
\def\thetabuh{\widehat{\thetabu}}

\def\fbh{\widehat{\fb}}
\def\Pbh{\widehat{\Pb}}
\def\Sigmabh{\widehat{\Sigmab}}
\def\Abh{\widehat{\Ab}}
\def\Bbh{\widehat{\Bb}}
\def\thetabh{\widehat{\thetab}}
\def\Rbe{\Rb_{\epsilon}}
\def\Rbeh{\widehat{\Rbe}}

\def\Rjk{{R_j}_k}
\def\fbik{{\fb_i}_k}
\def\fbjk{{\fb_j}_k}

\def\mik{{m_i}_k}
\def\sigmaik{{\sigma_i^2}_k}
\def\Sigmaik{{\Sigmab_i}_k}
\def\alphaik{{\alpha_i}_k}
\def\betaik{{\beta_i}_k}

\def\muik{{\mu_i}_k}
\def\vik{{v_i}_k}

\def\mjk{{m_j}_k}
\def\sigmajk{{\sigma_j^2}_k}
\def\Sigmajk{{\Sigmab_j}_k}
\def\alphajk{{\alpha_j}_k}
\def\betajk{{\beta_j}_k}

\def\mujk{{\mu_j}_k}
\def\vjk{{v_j}_k}
\def\njk{{n_j}_k}

\def\alphae{\alpha_{\epsilon}}
\def\betae{\beta_{\epsilon}}
\def\alphaez{\alpha_{\epsilon_0}}
\def\betaez{\beta_{\epsilon_0}}
\def\alphaei{\alpha_{\epsilon_i}}
\def\betaei{\beta_{\epsilon_i}}
\def\alphaeiz{\alpha_{\epsilon_{i_0}}}
\def\betaeiz{\beta_{\epsilon_{i-0}}}

\def\mbz{{{\mb}_z}}
\def\Sigmabz{{{\Sigmab}_z}}
\def\diag#1{\mbox{diag}\left[#1\right]}

\def\sigmah{\widehat{\sigma}}
\def\fh{\widehat{f}}
\def\sigmaz{\sigma_z}

\def\sumi{\sum_{i=1}^{M} }
\def\sumj{\sum_{j=1}^{N} }
\def\lambdah{\wh{\lambda}}
\def\lambdabh{\wh{\lambdab}}

\def\det#1{\hbox{det}\left(#1\right)}
\def\defined{\,\shortstack{$\triangle$\\ =}\,}

\def\zmat{\left[\matrix{0}\right]}
\def\det#1{\hbox{det}(#1)}
\def\th{\widehat{t}}
\def\xh{\widehat{x}}
\def\lambdah{\widehat{\lambda}}
\def\muh{\widehat{\mu}}
\def\sigmah{\widehat{\sigma}}
\def\rhoh{\widehat{\rho}}
\def\betah{\widehat{\beta}}
\def\alphah{\widehat{\alpha}}
\def\tauh{\widehat{\tau}}

\def\tbh{\widehat{\tb}}
\def\mubh{\widehat{\mub}}
\def\sigmabh{\widehat{\sigmab}}
\def\rhobh{\widehat{\rhob}}
\def\oneb{\mbox{\mathbold 1}}

\def\bm#1{\mbox{\boldmath $#1$}}
\def\zerob{{\mathbold 0}}
\def\oneb{{\mathbold 1}}
\def\intg{\int\kern-1.1em\int}
\def\expf#1{\exp\left[ {#1} \right]}

\def\gbu{\underline{\gb}}
\def\fbu{\underline{\fb}}
\def\zbu{\underline{\zb}}
\def\epsilonbu{\underline{\epsilonb}}
\def\uthetab{\underline{\thetab}}

\def\sigmae{{\sigma_{\epsilon}}}
\def\Sigmae{{\Sigma_{\epsilonb}}}
\def\Sigmabe{{\Sigmab_{\epsilonb}}}
\def\Sigmaf{{\Sigmab_{\fb}}}
\def\Sigmag{{\Sigmab_{\gb}}}

\def\fbuh{\widehat{\fbu}}
\def\zbuh{\widehat{\fbu}}
\def\uthetabh{\widehat{\uthetab}}

\def\fbh{\widehat{\fb}}
\def\Sigmabh{\widehat{\Sigmab}}
\def\Abh{\widehat{\Ab}}
\def\thetabh{\widehat{\thetab}}

\def\Rjk{{R_j}_k}
\def\fbik{{\fb_i}_k}
\def\fbjk{{\fb_j}_k}

\def\mik{{m_i}_k}
\def\sigmaik{{\sigma_i^2}_k}
\def\Sigmaik{{\Sigmab_i}_k}
\def\alphaik{{\alpha_i}_k}
\def\betaik{{\beta_i}_k}

\def\muik{{\mu_i}_k}
\def\vik{{v_i}_k}

\def\mjk{{m_j}_k}
\def\sigmajk{{\sigma_j^2}_k}
\def\Sigmajk{{\Sigmab_j}_k}
\def\alphajk{{\alpha_j}_k}
\def\betajk{{\beta_j}_k}

\def\mujk{{\mu_j}_k}
\def\vjk{{v_j}_k}
\def\njk{{n_j}_k}

\def\mbz{{{\mb}_z}}
\def\Sigmabz{{{\Sigmab}_z}}

\def\vect{\mbox{vect}}
\def\lra{\longrightarrow}

\def\gir{g_i(\rb)}
\def\fjr{f_j(\rb)}
\def\zjr{z_j(\rb)}
\def\epsilonir{\epsilon_i(\rb)}
\def\gbr{\gb(\rb)}
\def\fbr{\fb(\rb)}
\def\zbr{\zb(\rb)}
\def\epsilonbr{\epsilonb(\rb)}
\def\fbhr{\fbh(\rb)}
\def\Sigmabhr{\Sigmabh(\rb)}

\def\mbzr{\mb_{z(\rb)}}
\def\Sigmabzr{\Sigmab_{z(\rb)}}
\def\Sigmagbzr{{\Sigmab_g}_{z(\rb)}}

\def\mjzjr#1{{m_{#1}}_{z_{#1}(\rb)}}
\def\sigmajzjr#1{{\sigma_{#1}}_{z_{#1}(\rb)}}

\def\Beta{B}
\def\Betab{\bm{\Beta}}
\def\Betabe{\bm{\Beta}_{\epsilonb}}

\def\fh{\widehat{f}}
\def\sigmah{\widehat{\sigma}}
\def\sigmaei{\sigma_{\epsilon_i}^2}

\def\thetah{\widehat{\theta}}
\def\sigmah{\widehat{\sigma}}

\def\Ftxm{\widetilde{F}_{X}(x_-)}
\def\Ftxp{\widetilde{F}_{X}(x_+)}

\def\NU{{\cal V}}

\def\tFx{\widetilde{F}_{X}}

\def\Fx{F_{X}(x)}
\def\fx{f_{X}(x)}

\def\Fxv{F_{X}(x)}
\def\fxv{f_{X}(x)}

\def\Fxi{{F}_{X_i}(x_i)}
\def\fxi{{f}_{X_i}(x_i)}

\def\Gnu{G_{\NU}(\nu)}
\def\gnu{g_{\NU}(\nu)}

\def\Fnu{F_{\NU}(\nu)}
\def\fnu{f_{\NU}(\nu)}

\def\Finnu#1{F^{-1}_{\NU}(#1)}

\def\Gnua#1{G_{\NU}(#1)}
\def\Gnub#1{G_{\NU}^{-1}(#1)}
\def\Gnuh{G_{\NU}^{-1}(1/2)}

\def\Ftx{\widetilde{F}_{X}(x)}
\def\ftx{\widetilde{f}_{X}(x)}

\def\Ftxi{\widetilde{F}_{X_i}(x_i)}
\def\ftxi{\widetilde{f}_{X_i}(x_i)}

\def\Ftxv{\widetilde{F}_{X}(x)}
\def\FtX#1{\widetilde{F}_{X}(#1)}
\def\ftxv{\widetilde{f}_{X}(x)}

\def\fxnu{f_{X|\NU}(x|\nu)}
\def\fxNU{f_{X|\NU}(x|\NU)}
\def\FxNU{F_{X|\NU}(x|\NU)}

\def\fxNUtheta{f_{X|\NU,\theta}(x|\NU,\theta)}
\def\FxNUtheta{F_{X|\NU,\theta}(x|\NU,\theta)}

\def\Fxnu{F_{X|\NU}(x|\nu)}
\def\Fxnui#1{F_{X|\NU}(x|\nu_#1)}

\def\FxN#1{F_{X|\NU}(#1)}
\def\FxNU{F_{X|\NU}(x|\NU)}
\def\Fxnun#1{F_{X|\NU}(x|#1)}
\def\Fxinun#1{F_{X_i|\NU}(x_i|#1)}
\def\fxnu{f_{X|\NU}(x|\nu)}
\def\fxNU{f_{X|\NU}(x|\NU)}

\def\Fxvnu{F_{X|\NU}(x|\nu)}
\def\Fxvnun#1{F_{X|\NU}(x|#1)}
\def\FxvNU{F_{X|\NU}(x|\NU)}

\def\FXvnu{F_{X|\NU}}

\def\fxvnu{f_{X|\NU}(x|\nu)}
\def\fxvNU{f_{X|\NU}(x|\NU)}

\def\FXvNU#1{{F}_{X|\NU}(#1|\NU)}
\def\FXvn#1{{F}_{X|\NU}(x|#1)}

\def\Fxtheta{F_{X|\theta}(x|\theta)}
\def\fxtheta{f_{X|\theta}(x|\theta)}

\def\Fxvtheta{F_{X|\theta}(x|\theta)}
\def\fxvtheta{f_{X|\theta}(x|\theta)}

\def\Ftxtheta{\widetilde{F}_{X|\theta}(x|\theta)}
\def\ftxtheta{\widetilde{f}_{X|\theta}(x|\theta)}

\def\Ftxvtheta{\widetilde{F}_{X|\theta}(x|\theta)}
\def\ftxvtheta{\widetilde{f}_{X|\theta}(x|\theta)}

\def\Fxnutheta{F_{X|\NU,\theta}(x|\nu,\theta)}
\def\fxnutheta{f_{X|\NU,\theta}(x|\nu,\theta)}

\def\ftheta{f_{\Theta}(\theta)}
\def\fthetaxnuz{f_{\Theta|X,\nu_0}(\theta|x,\nu_0)}
\def\fthetax{f_{\Theta|X}(\theta|x)}

\def\Fxvnutheta{F_{X|\NU,\theta}(x|\nu,\theta)}
\def\FxvNUtheta{F_{X|\NU,\theta}(x|\NU,\theta)}
\def\fxvnutheta{f_{X|\NU,\theta}(x|\nu,\theta)}
\def\Ftxvnutheta{\widetilde{F}_{X|\theta}(x|\theta)}
\def\ftxvnutheta{\widetilde{f}_{X|\theta}(x|\theta)}

\def\FxvNUm{F_{X|\NU}(x_-|\NU)}
\def\FxvNUp{F_{X|\NU}(x_+|\NU)}

\def\Fxinu{F_{X_i|\NU}(x_i|\nu)}
\def\fxinu{f_{X_i|\NU}(x_i|\nu)}

\def\Ftxi{\widetilde{F}_{X_i}(x_i)}
\def\ftxi{\widetilde{f}_{X_i}(x_i)}

\def\fxitheta{f_{X_i|\theta}(x_i|\theta)}
\def\ftxitheta{\widetilde{f}_{X_i|\theta}(x_i|\theta)}

\def\thetah{\widehat{\theta}}
\def\thetat{\widetilde{\theta}}

\def\lnuxv{l_{\nu|X}(\nu|x)}

\def\Lnuxva#1{L_{\nu|X}(#1 | x)}
\def\lnuxva#1{l_{\nu|X}(#1 | x)}

\def\Prob#1{\mbox{P}\left(#1\right)}

\def\lthetaxvnuz{l(\theta)}

\def\lthetaxvnuz{l_{\theta|x,\nu_0}(\theta | x,\nu_0)}
\def\fxnuztheta{f_{X|\nu_0,\theta}(x | \nu_0,\theta)}
\def\fxinuztheta{f_{X|\nu_0,\theta}(x_i | \nu_0,\theta)}
\def\fxvtheta{f_{X|\theta}(x | \theta)}
\def\lthetaxv{l_{\theta|X}(\theta | x)}
\def\ltthetaxv{\widetilde{l}_{\theta|X}(\theta | x)}

\def\ltheta{l(\theta )}
\def\lttheta{\widetilde{l}(\theta)}
\def\Lnu{L(\nu )}
\def\Lnui#1{L(\nu_#1)}
\def\Linu{L_i(\nu )}
\def\Linu{L_i(\nu )}
\def\LNU{L(\NU )}
\def\Lin#1{L^{-1}(#1)}
\def\L#1{L(#1)}
\def\Li#1{L_i(#1)}
\def\FNU#1{F_\NU(#1)}
\def\FinNU#1{F^{-1}_\NU(#1)}

\def\ftthetax{\tilde{f}_{\Theta|X}(\theta|x)}
\def\ER{R}

\def\Ndd#1#2#3{{\cal N}\left( #1 ;~ #2, #3 \right)}
\def\Cdd#1#2#3{{\cal C}\left( #1 ;~ #2, #3 \right)}
\def\Sdd#1#2#3#4{{\cal S}\left( #1 ;~ #2, #3, #4 \right)}
\def\Edd#1#2{{\cal E}\left( #1 ;~ #2 \right)}
\def\DEdd#1#2{{\cal DE}\left( #1 ;~ #2 \right)}
\def\Gdd#1#2#3{{\cal G}\left( #1 ;~ #2, #3 \right)}
\def\IGdd#1#2#3{{\cal IG}\left( #1 ;~ #2, #3 \right)}

\def\Xwr{X(\omega,\rb)}
\def\xwr{x(\omega,\rb)}

\def\muzw{\mu_{z}(\omega)}
\def\muzwr{\mu_{z(\rb)}(\omega)}
\def\szw{\sigma_{z}^2(\omega)}
\def\szwr{\sigma_{z(\rb)}^2(\omega)}
\def\pzw{p_{z}(\omega)}
\def\pzwr{p_{z(\rb)}(\omega)}

\def\hszw{\widehat{\sigma_{z}^2}(\omega)}
\def\hmuzw{\widehat{\mu}_{z}(\omega)}
\def\hpzw{\widehat{p}_{z}(\omega)}

\def\muzwrl{\mu_{z(\rb)}(\omega-l)}

\def\pxwrcz{p\left(\xwr ~|~ z\right)}
\def\pxwr{p\left(\xwr\right)}

\def\Xbw{\Xb(\omega)}
\def\xbw{\xb(\omega)}
\def\uXb{\underline{\Xb}}
\def\uxb{\underline{\xb}}
\def\pxbw{p(\xbw)}
\def\puxb{p(\uxb)}
\def\pzb{P(\zb)}

\def\Szw{S_z(\omega)}
\def\mubw{\mub(\omega)}

\def\Srw{S_{\rb}(\omega)}
\def\Swr{S_{\omega}(\rb)}
\def\Sbw{\Sb(\omega)}
\def\sbw{\sb(\omega)}
\def\Sbr{\Sb(\rb)}
\def\mubw{\mub(\omega)}
\def\cov#1{\mbox{cov}[#1]}

\def\pdf{\emph{pdf}}
\def\pmf{\emph{pmf}}
\def\dpdx#1#2{\frac{\partial #1}{\partial #2}}

\def\REM#1{}

\def\XXwr{X(\omega,\rb)}
\def\xxwr{x(\omega,\rb)}

\def\Xwr{X_{\omega}(\rb)}
\def\xwr{x_{\omega}(\rb)}
\def\Xrw{X_{\rb}(\omega)}
\def\xrw{x_{\rb}(\omega)}

\def\Ywr{Y_{\omega}(\rb)}
\def\ywr{y_{\omega}(\rb)}
\def\Yrw{Y_{\rb}(\omega)}
\def\yrw{y_{\rb}(\omega)}

\def\Ewr{E_{\omega}(\rb)}
\def\ewr{\epsilon_{\omega}(\rb)}
\def\Erw{E_{\rb}(\omega)}
\def\erw{\epsilon_{\rb}(\omega)}

\def\Xbr{\Xb(\rb)}
\def\xbr{\xb(\rb)}
\def\Xbw{\Xb(\omega)}
\def\xbw{\xb(\omega)}

\def\uXb{\underline{\Xb}}
\def\uxb{\underline{\xb}}
\def\uXbz{\underline{\Xb}_z}
\def\uxbz{\underline{\xb}_z}

\def\uthetab{\underline{\thetab}}
\def\uthetabz{\underline{\thetab}_z}

\def\Ywr{Y_{\omega}(\rb)}
\def\ywr{y_{\omega}(\rb)}
\def\Yrw{Y_{\rb}(\omega)}
\def\yrw{y_{\rb}(\omega)}
\def\Ybr{\Yb(\rb)}
\def\ybr{\yb(\rb)}
\def\Ybw{\Yb(\omega)}
\def\ybw{\yb(\omega)}
\def\uYb{\underline{\Yb}}
\def\uyb{\underline{\yb}}
\def\uYbz{\underline{\Yb}_z}
\def\uybz{\underline{\yb}_z}

\def\Ewr{E_{\omega}(\rb)}
\def\ewr{\epsilon_{\omega}(\rb)}
\def\Erw{E_{\rb}(\omega)}
\def\erw{\epsilon_{\rb}(\omega)}
\def\Ebr{Eb(\rb)}
\def\ebr{\epsilonb(\rb)}
\def\Ebw{Eb(\omega)}
\def\ebw{\epsilonb(\omega)}
\def\uEb{\underline{\Eb}}

\def\Erwp{E_{\rb}(\omega')}
\def\Xrwp{X_{\rb}(\omega')}

\def\xwmr{x_{\omega-1}(\rb)}
\def\muzwm{\mu_{z}(\omega-1)}

\def\rhoz{\rho_z}
\def\sez{\sigma_{\eta_z}^2}

\def\xwrp{x_{\omega}(\rb')}
\def\xrwp{x_{\rb}(\omega')}
\def\xwpr{x_{\omega'}(\rb)}
\def\xwprp{x_{\omega'}(\rb')}
\def\Xwrp{x_{\omega}(\rb')}
\def\Xwrp{X_{\omega}(\rb')}
\def\Xwpr{X_{\omega'}(\rb)}
\def\Xwprp{X_{\omega'}(\rb')}

\def\sigmaed{\sigmae^2}
\def\sigmaew{\sigmae(\omega)}
\def\sigmaedw{\sigmaed(\omega)}

\def\Sigmabe{\Sigmab_{\epsilon}}
\def\sigmabh{\widehat{\Sigmab}}
\def\xbh{\widehat{\xb}}
\def\zbh{\widehat{\zb}}
\def\qbh{\widehat{\qb}}
\def\sbh{\widehat{\sb}}

\def\espx#1#2{\mbox{E}_{#1}\left\{#2\right\}}
\def\esp#1{\mbox{E}\left\{#1\right\}}
\def\d#1{\mbox{~d}#1}

\def\Tr#1{\mbox{Tr}\left[#1\right]}

\def\pmatrix#1{\left[\barr{ccccccccccccccc} #1 \earr\right]}
\def\Bf{\bm{B}}
\def\gbh{\widehat{\gb}}
\def\dbh{\widehat{\db}}
\def\Tr#1{\mbox{Tr}\left[#1\right]}
\def\Cov#1{\mbox{Cov}\left[#1\right]}
\def\iid{i.i.d.~}

\def\cro#1{\left[#1\right]}
\def\acc#1{\left\{#1\right\}}
\def\aprio{\emph{a priori~}}
\def\apost{\emph{a posteriori~}}
\def\hbh{\widehat{\hb}}
\def\qh{\widehat{q}}
\def\fbt{\widetilde{\fb}}
\def\thetabt{\widetilde{\thetab}}

\def\mubt{\widetilde{\mub}}
\def\Sigmabt{\widetilde{\Sigmab}}
\def\qt{\widetilde{q}}
\def\vt{\widetilde{v}}
\def\zt{\widetilde{z}}
\def\Dbt{\widetilde{\Db}}
\def\alphat{\widetilde{\alpha}}
\def\betat{\widetilde{\beta}}
\def\abt{\widetilde{\ab}}
\def\ajt{\widetilde{a}_j}
\def\taut{\widetilde{\tau}}
\def\mut{\widetilde{\mu}}
\def\Zbt{\widetilde{\Zb}}
\def\Abt{\widetilde{\Ab}}
\def\zbt{\widetilde{\zb}}

\def\bloca#1#2#3{
\begin{picture}(50,50)
  \put(32,42){\makebox(0,0){#1}}
  \put(25,50){\vector(0,-1){20}}
  \put(0,10){\framebox(50,20){#2}}
  \put(0,0){\makebox(50,0){#3}}  
\end{picture}
}

\def\ugb{\underline{\gb}}
\def\fbh{\widehat{\fb}}
\def\zbh{\widehat{\zb}}
\def\qbh{\widehat{\qb}}
\def\thetabh{\widehat{\thetab}}
\def\Pbh{\widehat{\Pb}}

\def\sigmae{\sigma_{\epsilon}}
\def\REM#1{}
\def\fh{\widehat{f}}
\def\zh{\widehat{z}}
\def\disp{\displaystyle}
\def\oneb{{\boldmath{1}}}
\def\zerob{{\boldmath{0}}}
\def\d#1{\; \mbox{d} #1}
\def\expf#1{\exp\left\{#1\right\}}
\def\esp#1{\mbox{E}\left\{#1\right\}}
\def\lra{\longrightarrow}

\def\bslide#1{\begin{frame}{\Large\mathbold #1}}
\def\eslide{\end{frame}}

\def\inversions#1#2#3{
\begin{picture}(100,40)
  \put(0,15){\makebox(15,0){#1}}
  \put(20,15){\vector(1,0){10}}
  \put(30,7){\framebox(40,15){#2}}
  \put(70,15){\vector(1,0){10}}
  \put(85,15){\makebox(15,0){#3}}
\end{picture}
}

\def\rfb{\red{\fb}}
\def\bgb{\blue{\gb}}
\def\rfbh{\red{\widehat{\fb}}}
\def\rfbt{\red{\widetilde{\fb}}}
\def\rzbt{\red{\widetilde{\zb}}}
\def\rqb{\red{\qb}}
\def\rqbh{\red{\widehat{\qb}}}
\def\rzb{\red{\zb}}
\def\rzbh{\red{\widehat{\zb}}}
\def\rthetab{\red{\thetab}}
\def\retab{\red{\etab}}
\def\rthetabh{\red{\widehat{\thetab}}}
\def\rthetabt{\red{\widetilde{\thetab}}}
\def\retabt{\red{\widetilde{\etab}}}
\def\ralphabt{\red{\widetilde{\alphab}}}

\def\ugb{\underline{\gb}}
\def\fbh{\widehat{\fb}}
\def\zbh{\widehat{\zb}}
\def\qbh{\widehat{\qb}}
\def\thetabh{\widehat{\thetab}}
\def\Pbh{\widehat{\Pb}}

\def\UP#1{\uppercase{#1}}
\def\sca#1{\uppercase{#1}}

\def\sigmae{\sigma_{\epsilon}}
\def\REM#1{}
\def\fh{\widehat{f}}
\def\zh{\widehat{z}}
\def\disp{\displaystyle}
\def\oneb{{\boldmath{1}}}
\def\zerob{{\boldmath{0}}}
\def\d#1{\; \mbox{d} #1}
\def\expf#1{\exp\left\{#1\right\}}
\def\esp#1{\mbox{E}\left\{#1\right\}}
\def\lra{\longrightarrow}

\def\rfb{\red{\fb}}
\def\rFb{\red{\Fb}}
\def\bgb{\blue{\gb}}
\def\rfbh{\red{\widehat{\fb}}}
\def\rqb{\red{\qb}}
\def\rqbh{\red{\widehat{\qb}}}
\def\rzb{\red{\zb}}
\def\rzbh{\red{\widehat{\zb}}}
\def\rthetab{\red{\thetab}}
\def\ralphab{\red{\alphab}}
\def\rbetab{\red{\betab}}
\def\rthetabh{\red{\widehat{\thetab}}}
\def\ralphabh{\red{\widehat{\alphaab}}}
\def\rSigmab{\red{\Sigmab}}
\def\rSigmabe{\red{\Sigmabe}}

\def\rz{\red{z}}
\def\rzh{\red{\widehat{z}}}

\def\rab{\red{\ab}}
\def\rabh{\red{\widehat{\ab}}}
\def\rAb{\red{\Ab}}
\def\rAbh{\red{\widehat{\Ab}}}
\def\rBb{\red{\Bb}}
\def\rBbh{\red{\widehat{\Bb}}}
\def\rPbh{\red{\widehat{\Pb}}}
\def\rub{\red{\ub}}
\def\rubh{\red{\widehat{\ub}}}

\def\rVb{\red{\Vb}}
\def\rVbh{\widehat{\rVb}}
\def\rVbt{\widetilde{\rVb}}

\def\rvb{\red{\vb}}
\def\rvbh{\widehat{\rvb}}
\def\rvbt{\widetilde{\rvb}}

\def\bfb{\blue{\fb}}
\def\bFb{\blue{\Fb}}

\def\rfi{\red{f}_i}
\def\rfj{\red{f}_j}
\def\ruj{\red{u}_j}
\def\raj{\red{a}_j}
\def\rzj{\red{z}_j}
\def\rvj{\red{v}_j}
\def\rtau{\red{\tau}}
\def\rtauj{\red{\tau_j}}
\def\rtaub{\red{\taub}}
\def\rtauh{\widehat{\rtau}}
\def\rtaujh{\widehat{\rtauj}}
\def\rtaubh{\widehat{\rtaub}}
\def\rve{\red{v_{\epsilon}}}
\def\rv{\red{v}}

\def\rfjh{\red{\widehat{f}}_j}
\def\rujh{\red{\widehat{u}}_j}
\def\rajh{\red{\widehat{a}}_j}
\def\rzjh{\red{\widehat{z}}_j}
\def\rvjh{\red{\widehat{v}}_j}

\def\ralpha{\red{\alpha}}
\def\rbeta{\red{\beta}}
\def\rtheta{\red{\theta}}

\def\rlambda{\red{\lambda}}
\def\ralphah{\widehat{\ralpha}}
\def\rbetah{\widehat{\rbeta}}
\def\rthetah{\widehat{\rtheta}}
\def\rlambdah{\widehat{\rlambda}}

\def\ubh{\widehat{\ub}}

\def\rve{\red{v_{\epsilon}}}
\def\alphae{\alpha_{\epsilon_0}}
\def\betae{\beta_{\epsilon_0}}
\def\alphajh{\widehat{\alpha}_j}
\def\betajh{\widehat{\beta}_j}
\def\alphaeh{\widehat{\alpha}_{\epsilon_0}}
\def\betaeh{\widehat{\beta}_{\epsilon_0}}

\def\rqb{\red{\qb}}
\def\rqbh{\red{\widehat{\qb}}}
\def\rzb{\red{\zb}}
\def\rZb{\red{\Zb}}
\def\rab{\red{\ab}}
\def\rAb{\red{\Ab}}
\def\rvb{\red{\vb}}
\def\rmb{\red{\mb}}
\def\rzbh{\red{\widehat{\zb}}}
\def\rvbh{\red{\widehat{\vb}}}
\def\rmbh{\red{\widehat{\mb}}}
\def\rthetab{\red{\thetab}}
\def\rthetabh{\red{\widehat{\thetab}}}
\def\rfbh{\widehat{\rfb}}
\def\rhb{\red{\hb}}
\def\rhbh{\widehat{\rhb}}
\def\rthetai{\red{\theta}_i}
\def\rthetab{\red{\thetab}}
\def\rfjm{\red{f}_{j-1}}
\def\rfim{\red{f}_{i-1}}

\def\muh{\widehat{\mu}}
\def\vh{\widehat{v}}

\def\rSigmabh{\red{\Sigmabh}}
\def\rVbh{\red{\Vbh}}
\def\Fbh{\red{\Fb}}
\def\rFbh{\red{\Fbh}}

\def\Sigmabeh{\widehat{\Sigmabe}}
\def\rSigmabeh{\widehat{\red{\Sigmabe}}}

\def\rf{\red{f}}
\def\bg{\blue{g}}
\def\rF{\red{F}}
\def\bG{\blue{G}}

\def\sumk{\sum_{k=1}^K}
\def\suml{\sum_{l=1}^L}
\def\sumn{\sum_{n=1}^N}
\def\sumj{\sum_{j=1}^J}
\def\sumi{\sum_{i=1}^I}

\def\nub{\bm{\nu}}
\def\rak{\red{a_k}}
\def\rbk{\red{b_k}}
\def\rck{\red{c_k}}
\def\rtk{\red{t_k}}
\def\rvk{\red{v_k}}
\def\rnuk{\red{\nu_k}}
\def\rnuz{\red{\nu_0}}
\def\rphik{\red{\phi_k}}
\def\rnub{\red{\nub}}
\def\rtb{\red{\tb}}
\def\rphib{\red{\phib}}
\def\rfn{\red{f_n}}
\def\rtn{\red{t_n}}
\def\rvn{\red{v_n}}

\def\rve{\red{v_\epsilon}}
\def\rvf{\red{v_f}}
\def\rveh{\red{\widehat{v}_\epsilon}}
\def\rvfh{\red{\widehat{v}_f}}

\def\rnu{\red{\nu}}
\def\rphi{\red{\phi}}
\def\rbl{\red{b_l}}
\def\rbb{\red{\bb}}

\def\modela{\tt\setlength{\unitlength}{0.92pt}
\begin{picture}(250,106)
\thinlines    \put(-5,105){$\Nc(t|\red{t_1,v_1})$}
              \put(35,100){\vector(1,0){25}}
              \put(60,95){\framebox(37,16){\red{$a_1$}}}
			  \multiput(77,72)(0,7){3}{\circle*{4}}
              \put(-5,65){$\Nc(t|\red{t_k,v_k})$}
              \put(35,59){\vector(1,0){25}}
              \put(60,51){\framebox(37,16){\red{$a_k$}}}
              \multiput(77,32)(0,7){3}{\circle*{4}}              
              \put(-5,23){$\Nc(t|\red{t_K,v_K})$}
              \put(60,10){\framebox(37,16){\red{$a_K$}}}
              \put(35,18){\vector(1,0){25}}

              \put(98,103){\vector(1,-1){38}}
              \put(98,59){\vector(1,0){39}}
              \put(98,17){\vector(1,1){38}}

              \put(150,60){\circle{24}}
              \put(146,57){$\Large+$}
              \put(165,75){$g_{\rthetab}(t)$}
              \put(164,60){\vector(1,0){33}}
              
              \put(212,85){$\epsilon(t)$}
              \put(210,95){\vector(0,-1){20}}
              \put(211,60){\circle{24}}
              \put(207,58){$\Large+$}
              
              \put(225,66){$g(t)$}
              \put(225,60){\vector(1,0){20}}
\end{picture}
}

\def\modelb{\tt\setlength{\unitlength}{0.92pt}
\begin{picture}(250,106)
\thinlines    \put(-5,105){$\delta(t-\red{t_1})$}
              \put(0,100){\vector(1,0){37}}
              \put(38,95){\framebox(60,16){$\red{a_1} \Nc(t|0,\red{v_1})$}}
			  \multiput(77,72)(0,7){3}{\circle*{4}}
              \put(-5,65){$\delta(t-\red{t_k})$}
              \put(0,59){\vector(1,0){37}}
              \put(38,51){\framebox(60,16){$\red{a_k} \Nc(t|0,\red{v_k})$}}
              \multiput(77,32)(0,7){3}{\circle*{4}}              
              \put(-5,23){$\delta(t-\red{t_K})$}
              \put(38,10){\framebox(60,16){$\red{a_K} \Nc(t|0,\red{v_K})$}}
              \put(0,18){\vector(1,0){37}}

              \put(98,103){\vector(1,-1){38}}
              \put(98,59){\vector(1,0){39}}
              \put(98,17){\vector(1,1){38}}

              \put(150,60){\circle{24}}
              \put(146,57){$\Large+$}
              \put(165,75){$g_{\rthetab}(t)$}
              \put(164,60){\vector(1,0){33}}
              
              \put(212,85){$\epsilon(t)$}
              \put(210,95){\vector(0,-1){20}}
              \put(211,60){\circle{24}}
              \put(207,58){$\Large+$}
              
              \put(225,66){$g(t)$}
              \put(225,60){\vector(1,0){20}}
\end{picture}
}

\def\modelc{\tt\setlength{\unitlength}{0.92pt}
\begin{picture}(250,106)
\thinlines    \put(-5,105){$\delta(t-\red{t_1})$}
              \put(0,100){\vector(1,0){37}}
              \put(38,95){\framebox(40,16){$\red{a_1}$}}
			  \multiput(57,72)(0,7){3}{\circle*{4}}
              \put(-5,65){$\delta(t-\red{t_k})$}
              \put(0,59){\vector(1,0){37}}
              \put(38,51){\framebox(40,16){$\red{a_k}$}}
              \multiput(57,32)(0,7){3}{\circle*{4}}              
              \put(-5,23){$\delta(t-\red{t_K})$}
              \put(38,10){\framebox(40,16){$\red{a_K}$}}
              \put(0,18){\vector(1,0){37}}

              \put(78,103){\vector(1,-1){38}}
              \put(78,59){\vector(1,0){39}}
              \put(78,17){\vector(1,1){38}}

              \put(130,60){\circle{24}}
              \put(126,57){$\Large+$}
              \put(144,60){\vector(1,0){10}}
              \put(154,52){\framebox(50,16){$\Nc(t|0,v)$}}
              
              \put(205,75){$g_{\rthetab}(t)$}
              \put(204,60){\vector(1,0){33}}
              
              \put(252,85){$\epsilon(t)$}
              \put(250,95){\vector(0,-1){20}}
              \put(251,60){\circle{24}}
              \put(247,58){$\Large+$}
              
              \put(265,66){$g(t)$}
              \put(265,60){\vector(1,0){20}}
\end{picture}
}

\def\modeld{\tt\setlength{\unitlength}{0.92pt}
\begin{picture}(250,106)
\thinlines    \put(20,70){$s(t)=\sum_k \rak \delta(t-\red{t_k})$}
              \put(105,60){\vector(1,0){50}}
              \put(154,52){\framebox(50,16){$\Nc(t|0,v)$}}
              
              \put(205,75){$g_{\rthetab}(t)$}
              \put(204,60){\vector(1,0){33}}
              
              \put(252,85){$\epsilon(t)$}
              \put(250,95){\vector(0,-1){20}}
              \put(251,60){\circle{24}}
              \put(247,58){$\Large+$}
              
              \put(265,66){$g(t)$}
              \put(265,60){\vector(1,0){20}}
              \put(0,20){\siga}
\end{picture}
}

\def\siga{\setlength{\unitlength}{0.46pt}
\begin{picture}(290,108)
\thinlines    \put(10,26){\vector(1,0){260}}
              \put(16,12){\vector(0,1){88}}
              \put(48,27){\vector(0,1){34}}
              \put(79,27){\vector(0,1){54}}
              \put(130,27){\vector(0,1){46}}
              \put(188,26){\vector(0,1){40}}
              \put(262,14){$t$}
              \put(182,16){$\red{t_K}$}
              \put(195,63){$\red{a_K}$}
              \put(125,15){$\red{t_k}$}
              \put(137,68){$\red{a_k}$}
              \put(75,14){$\red{t_2}$}
              \put(84,73){$\red{a_2}$}
              \put(43,15){$\red{t_1}$}
              \put(54,56){$\red{a_1}$}
              \multiput(150,46)(8,0){3}{\circle*{4}}
\end{picture}
}

\def\modele{\tt\setlength{\unitlength}{0.92pt}
\begin{picture}(250,106)
\thinlines    \put(20,70){$s(t)=\sum_n \rfn \delta(t-n\Delta t)$}
              \put(105,60){\vector(1,0){50}}
              \put(154,52){\framebox(50,16){$\Nc(t|0,v)$}}
              
              \put(205,75){$g_{\rfb}(t)$}
              \put(204,60){\vector(1,0){33}}
              
              \put(252,85){$\epsilon(t)$}
              \put(250,95){\vector(0,-1){20}}
              \put(251,60){\circle{24}}
              \put(247,58){$\Large+$}
              
              \put(265,66){$g(t)$}
              \put(265,60){\vector(1,0){20}}
              \put(0,20){\sigb}
\end{picture}
}

\def\sigb{\setlength{\unitlength}{0.46pt}
\begin{picture}(290,108)
\thinlines    \put(10,26){\vector(1,0){260}}
              \put(10,12){\vector(0,1){88}}
              
              \put(20,27){\vector(0,1){34}}
              \put(40,27){\vector(0,1){14}}
              \put(60,27){\vector(0,1){54}}
              \put(80,27){\vector(0,1){31}}
              \put(100,27){\vector(0,1){46}}
              \put(120,27){\vector(0,1){40}}
              \put(140,27){\vector(0,1){48}}
              \put(160,27){\vector(0,1){34}}
              \put(180,27){\vector(0,1){14}}
              \put(200,27){\vector(0,1){31}}
              \put(220,27){\vector(0,1){46}}
              \put(240,27){\vector(0,1){40}}
              
              \put(270,14){$t$}
              \put(15,10){$1$}
              \put(15,63){$\red{f_1}$}
              \put(135,10){$n$}
              \put(135,68){$\red{f_n}$}
              \put(235,10){$N$}
              \put(235,68){$\red{f_N}$}

\end{picture}
}

\def\sigbr{\setlength{\unitlength}{0.46pt}
\begin{picture}(290,108)
\thinlines    \put(10,26){\vector(1,0){260}}
              \put(10,12){\vector(0,1){88}}
              
              \put(20,27){\vector(0,1){34}}
              \put(40,27){\vector(0,1){14}}
              \put(60,27){\vector(0,1){54}}
              \put(80,27){\vector(0,1){31}}
              \put(100,27){\vector(0,1){46}}
              \put(120,27){\vector(0,1){40}}
              \put(140,27){\vector(0,1){48}}
              \put(160,27){\vector(0,1){34}}
              \put(180,27){\vector(0,1){14}}
              \put(200,27){\vector(0,1){31}}
              \put(220,27){\vector(0,1){46}}
              \put(240,27){\vector(0,1){40}}
              \put(270,14){$t$}
              \put(15,10){$1$}
              \put(15,63){$\red{f_1}$}
              \put(135,10){$n$}
              \put(135,68){$\red{f_n}$}
              \put(235,10){$N$}
              \put(235,68){$\red{f_N}$}

              \put(60,10){$\red{\Delta t}$}
\end{picture}
}

\def\modelf{\tt\setlength{\unitlength}{0.92pt}
\begin{picture}(250,106)
\thinlines    \put(20,70){$s(t)=\sum_n \rfn \delta(t-n\red{\Delta t})$}
              \put(105,60){\vector(1,0){50}}
              \put(154,52){\framebox(50,16){$\Nc(t|0,\red{v})$}}
              
              \put(205,75){$g_{\rfb,\rthetab}(t)$}
              \put(204,60){\vector(1,0){33}}
              
              \put(252,85){$\epsilon(t)$}
              \put(250,95){\vector(0,-1){20}}
              \put(251,60){\circle{24}}
              \put(247,58){$\Large+$}
              
              \put(265,66){$g(t)$}
              \put(265,60){\vector(1,0){20}}
              \put(0,20){\sigbr}
\end{picture}
}

\def\sig{\tt\setlength{\unitlength}{1.38pt}
\begin{picture}(290,108)
\thinlines    
              \put(10,26){\vector(1,0){260}}
              \put(48,27){\vector(0,1){34}}
              \put(16,12){\vector(0,1){88}}
              \put(79,27){\vector(0,1){54}}
              \put(102,27){\vector(0,1){31}}
              \put(130,27){\vector(0,1){46}}
              \put(188,26){\vector(0,1){40}}
              \put(262,14){$t$}
              \put(22,90){$g(t)$}
              \put(182,16){$t_K$}
              \put(195,63){$a_K$}
              \put(125,15){$t_k$}
              \put(137,68){$a_k$}
              \put(75,14){$t_2$}
              \put(84,73){$a_2$}
              \put(43,15){$t_1$}
              \put(54,56){$a_1$}
              \multiput(150,46)(8,0){3}{\circle*{4}}
\end{picture}
}

\def\decgb{\tt\setlength{\unitlength}{0.92pt}
\begin{picture}(363,122)
\thinlines    \put(248,66){\vector(1,0){14}}
              \put(215,57){\framebox(33,18){$h(t)$}}
              \put(288,66){\vector(1,0){22}}
              \put(270,64){$\large\Sigma$}
              \put(275,67){\circle{26}}
              \put(275,102){\vector(0,-1){20}}
              \put(280,94){$b(t)$}
              \put(296,75){$y(t)$}
              \multiput(100,35)(0,8){3}{\circle*{4}}
              \put(194,72){$x(t)$}
              \put(193,66){\vector(1,0){23}}
              \put(176,61){$\large\Sigma$}
              \put(181,65){\circle{26}}
              \put(80,92){\framebox(43,18){$a_1$}}
              \put(80,58){\framebox(43,18){$a_2$}}
              \put(79,10){\framebox(43,18){$a_K$}}
              \put(123,66){\vector(1,0){45}}
              \put(123,101){\vector(4,-3){44}}
              \put(122,18){\vector(1,1){44}}
              \put(55,101){\vector(1,0){23}}
              \put(55,66){\vector(1,0){23}}
              \put(55,19){\vector(1,0){23}}
              \put(10,104){$g_1(t-t_1)$}
              \put(10,69){$g_2(t-t_2)$}
              \put(10,22){$g_K(t-t_k)$}
\end{picture}
}

\def\decgc{\tt\setlength{\unitlength}{0.92pt}
\begin{picture}(287,68)
\thinlines    \put(36,20){\vector(1,0){23}}
              \put(101,20){\vector(1,0){23}}
              \put(167,20){\vector(1,0){16}}
              \put(124,12){\framebox(43,18){$h(t)$}}
              \put(210,20){\vector(1,0){23}}
              \put(192,20){$\large\Sigma$}
              \put(196,23){\circle{26}}
              \put(196,57){\vector(0,-1){20}}
              \put(200,49){$b(t)$}
              \put(218,30){$y(t)$}
              \put(15,20){$s(t)$}
              \put(101,32){$x(t)$}
              \put(59,11){\framebox(43,18){$g(t)$}}
\end{picture}
}

\def\sigd{
\begin{picture}(290,108)
\thinlines    \put(10,26){\vector(1,0){260}}
              \put(48,27){\vector(0,1){34}}
              \put(16,12){\vector(0,1){88}}
              \put(79,27){\vector(0,1){54}}
              \put(102,27){\vector(0,1){31}}
              \put(130,27){\vector(0,1){46}}
              \put(188,26){\vector(0,1){40}}
              \put(262,14){$t$}
              \put(22,90){$g(t)$}
              \put(182,16){$t_K$}
              \put(195,63){$a_K$}
              \put(125,15){$t_k$}
              \put(137,68){$a_k$}
              \put(75,14){$t_2$}
              \put(84,73){$a_2$}
              \put(43,15){$t_1$}
              \put(54,56){$a_1$}
              \multiput(150,46)(8,0){3}{\circle*{4}}
\end{picture}
}

\def\wt#1{\widetilde{#1}}
\def\Thetab{\bm{\Theta}}
\def\Thetabh{\widehat{\Thetab}}
\def\abh{\widehat{\ab}}
\def\kbt{\tilde{\kb}}
\def\kt{\tilde{k}}
\def\gammat{\tilde{\gamma}}
\def\Sigmat{\tilde{\Sigma}}
\def\etat{\tilde{\eta}}

\def\rHb{\red{\Hb}}
\def\rFb{\red{\Fb}}

\def\wt#1{\widetilde{#1}}
\def\Thetab{\bm{\Theta}}
\def\Thetabh{\widehat{\Thetab}}
\def\abh{\widehat{\ab}}
\def\kbt{\tilde{\kb}}
\def\kt{\tilde{k}}
\def\gammat{\tilde{\gamma}}
\def\Sigmat{\tilde{\Sigma}}
\def\etat{\tilde{\eta}}
\def\zetat{\tilde{\zeta}}
\def\Mt{\tilde{M}}
\def\Mh{\hat{M}}
\def\Nt{\tilde{N}}
\def\atk{{\tilde{a}_k}}
\def\ztk{{\tilde{z}_k}}

\def\rh{\red{h}}
\def\rvbe{\rvb_{\epsilon}}
\def\rvei{\rv_{\epsilon_i}}
\def\rVbe{\rVb_{\epsilon}}
\def\rVbeh{\red{\Vbh}_{\epsilon}}
\def\vbh{\hat{\vb}}
\def\rVbfh{\red{\Vbh}_f}
\def\rvbeh{\rvbh_{\epsilon}}
\def\rvbet{\rvbt_{\epsilon}}
\def\rvbfh{\rvbh_f}

\def\rvbf{\rvb_f}
\def\rvfj{\rv_{f_j}}
\def\rVbf{\red{\Vb}_f}

\def\alphaez{\alpha_{\epsilon_0}}
\def\betaez{\beta_{\epsilon_0}}
\def\alphafz{\alpha_{f_0}}
\def\betafz{\beta_{f_0}}

\def\wh#1{\widehat{#1}}
\def\disp#1{{\displaystyle #1}}

\def\ER{\mbox{I\kern-.25em R}}
\def\EC{\mbox{C\kern-.8em C}}
\def\EZ{\mbox{Z\kern-.55em Z}}
\def\EN{\mbox{N\kern-.8em N}}

\def\bm#1{\mbox{\boldmath $#1$}}
\def\sb{{\bm s}}
\def\ub{{\bm u}}
\def\xb{{\bm x}}
\def\yb{{\bm y}}
\def\db{{\bm{d}}}
\def\Ab{{\bm A}}
\def\Bb{{\bm B}}
\def\Fb{{\bm F}}
\def\Rb{{\bm R}}
\def\Sb{{\bm S}}
\def\Ub{{\bm U}}

\def\thetab{\bm{\theta}}
\def\phib{\bm{\phi}}
\def\lambdab{\bm{\lambda}}
\def\Lambdab{\bm{\Lambda}}
\def\Sigmab{\bm{\Sigma}}
\def\lra{\longrightarrow}
\def\intg{\int\kern-1.1em\int}
\def\intd{\int\kern-.8em\int}

\def\apriori{{\em a priori} }
\def\aposteriori{{\em a posteriori} }

\def\d#1{\,\mbox{d}#1}
\def\esp#1{\mbox{E}\left\{ #1 \right\}}
\def\espx#1#2{\mbox{E}_{#1}\left\{ #2 \right\}}
\def\expf#1{\exp\left[ {#1} \right]}
\def\dpdx#1#2{\frac{\partial #1}{\partial #2}}
\def\dpdxd#1#2{\frac{\partial^2 #1}{\partial #2^2}}

\def\ent#1{\hbox{H}\left[#1\right]}
\def\dent#1{\delta\hbox{H}\left[#1\right]}
\def\rent#1{\hbox{D}\left[#1\right]}
\def\kullback#1{\hbox{K}\left[#1\right]}
\def\pent#1{\hbox{PE}\left[#1\right]}
\def\mdiv#1{\hbox{J}\left[#1\right]}
\def\infmut#1{\hbox{I}\left[#1\right]}
\def\idisc#1{\hbox{ID}\left[#1\right]}
\def\j{\mbox{j}}

\def\pxtheta{p(x | \thetab)}
\def\pxthetas{p(x | \thetab^*)}
\def\pxthetasd{p(x | \thetab^*+\Delta\thetab)}

\def\tr#1{\hbox{tr}\left(#1\right)}
\def\det#1{\hbox{det}\left(#1\right)}

\def\ieeeSSC{IEEE Transactions on Systems Science and Cybernetics}
\def\ieeeSP{IEEE Transactions on Signal Processing}			
\def\ieeeP{Proceedings of the IEEE}					
\def\ieeeIT{IEEE Transactions on Information Theory}			
\def\TS{Traitement du Signal}						
\def\siamCO{SIAM Journal of Control}					
\def\AISM{Annals of Institute of Statistical Mathematics}		
\def\ieeeASSP{IEEE Transactions on Acoustics Speech and Signal Processing}
\def\ieeeAC{IEEE Transactions on Automatic and Control}			

\def\jan{janvier\xspace}
\def\feb{f\'evrier\xspace}
\def\mar{mars\xspace}
\def\apr{avril\xspace}
\def\may{mai\xspace}
\def\jun{juin\xspace}
\def\jul{juillet\xspace}
\def\aug{ao\^ut\xspace}
\def\sep{septembre\xspace}
\def\oct{octobre\xspace}
\def\nov{novembre\xspace}
\def\dec{d\'ecembre\xspace}
\def\Jan{January\xspace}	
\def\Feb{February\xspace}
\def\Mar{March\xspace}
\def\Apr{April\xspace}
\def\May{May\xspace}
\def\Jun{June\xspace}
\def\Jul{July\xspace}
\def\Aug{August\xspace}
\def\Sep{September\xspace}
\def\Oct{October\xspace}
\def\Nov{November\xspace}
\def\Dec{December\xspace}

\def\espq#1{\left<#1\right>_q}

\def\rZbe{\rZb_{\epsilon}}
\def\nubt{\widetilde{\nub}}

\def\rVbf{\rVb_{f}}
\def\rvbf{\rvb_{f}}
\def\rvfj{\red{v}_{f_j}}
\def\rvbhe{\rvbh_{\epsilon}}
\def\rvbhf{\rvbh_{f}}
\def\bsplit{\begin{split}}
\def\esplit{\end{split}}
\def\alphaei{\alpha_{\epsilon_i}}
\def\alphafj{\alpha_{f_j}}
\def\betaei{\beta_{\epsilon_i}}
\def\betafj{\beta_{f_j}}

\def\bgbh{\widehat{\bgb}}
\def\rfh{\widehat{\rf}}
\def\rFh{\widehat{\rF}}
\def\rc{\red{r}}
\def\rch{\widehat{\rc}}
\def\partialx{\frac{\partial }{\partial x}}
\def\partialy{\frac{\partial }{\partial y}}
\def\partialr{\frac{\partial }{\partial r}}
\def\partialxd{\frac{\partial^2 }{\partial x^2}}
\def\partialyd{\frac{\partial^2 }{\partial y^2}}
\def\partialrd{\frac{\partial^2 }{\partial r^2}}
\def\cosphi{\cos(\phi)}
\def\sinphi{\sin(\phi)}
\def\sumk{\sum_{k=1}^K}
\def\sumj{\sum_{j=1}^N}
\def\sumi{\sum_{i=1}^M}
\def\summ{\sum_{m=1}^M}
\def\sumn{\sum_{n=1}^N}
\def\summn{\sum_{m=1}^M\sum_{n=1}^N}

\def\df{\dot{f}}
\def\ddf{\ddot{f}}

\def\rf{\red{f}}
\def\rdf{\red{\df}}
\def\rddf{\red{\ddf}}

\def\rfb{\red{\fb}}
\def\rdfb{\red{\dot{\fb}}}
\def\rddfb{\red{\ddot{\fb}}}

\def\bdg{\blue{\dot{g}}}
\def\bddg{\blue{\ddot{g}}}

\def\bgb{\blue{\gb}}
\def\bdg{\blue{\dot{g}}}

\def\bdgb{\blue{\dot{\gb}}}
\def\bddgb{\blue{\ddot{\gb}}}

\def\epsilonbd{\dot{\epsilonb}}
\def\epsilonbdd{\ddot{\epsilonb}}

\def\rdfbh{\widehat{\rdfb}}
\def\rddfbh{\widehat{\rddfb}}

\def\rqh{\red{\widehat{q}}}
\def\rdfh{\red{\widehat{\dot{f}}}}
\def\rddfh{\red{\widehat{\ddot{f}}}}

\def\vek{v_{\epsilon_k}}
\def\rveih{\widehat{\rve}_i}
\def\rvfjh{\widehat{\rvf}_j}

\def\alphaxiz{{\alpha_{\xi_0}}}
\def\betaxiz{{\beta_{\xi_0}}}
\def\alphazz{{\alpha_{z_0}}}

\def\betazz{{\beta_{z_0}}}
\def\rVbz{{\rVb_z}}
\def\rvbz{{\rvb_z}}
\def\rvxi{{\rv_\xi}}
\def\rvzj{{\rv_{z_j}}}
\def\rvxih{{\widehat{\rv}_\xi}}
\def\rvbzh{{\widehat{\rvb}_z}}
\def\rvzjh{{\widehat{\rv}_{z_j}}}

\def\alphae{{\alpha_\epsilon}}

\def\alphaet{{\widetilde{\alpha}_\epsilon}}
\def\alphaft{{\widetilde{\alpha}_f}}
\def\betae{{\beta_\epsilon}}
\def\betaet{{\widetilde{\beta}_\epsilon}}
\def\betaft{{\widetilde{\beta}_f}}
\def\rfih{\hat{\rf}_i}
\def\rfbk{\rfb^{(k)}}
\def\rfbkp{\rfb^{(k+1)}}

\def\rmh{\hat{\red{m}}}

\def\rmk{{\red{m}_k}}
\def\rmkh{{\hat{\red{m}}_k}}
\def\rvk{{\red{v}_k}}
\def\rvkh{{\hat{\red{v}}_k}}
\def\rak{{\red{a}_k}}
\def\rakh{{\hat{\red{a}}_k}}

\def\rvh{\hat{\red{v}}}
\def\rveh{\hat{\rve}}

\def\rmt{\tilde{\red{m}}}
\def\rvt{\tilde{\red{v}}}
\def\rvet{\tilde{\rve}}
\def\rveih{\red{\hat{v}}_{\epsilon_i}}

\def\rveb{\red{\vb_{\epsilon}}}
\def\rvebh{\red{\vbh_{\epsilon}}}

\def\blocgi{\begin{picture}(80,50)
  \put(60,40){\makebox(0,0){$\red{m}$}}
  \put(60,40){\circle{20}}
  \put(60,30){\vector(0,-1){20}}
  \put(60,0){\circle{20}}
  \put(60,0){\makebox(0,0){$\blue{g}_i$}}
  \put(20,0){\circle{20}}
  \put(20,0){\makebox(0,0){$\epsilon_i$}}
  \put(30,0){\vector(1,0){20}}
\end{picture}
}
\def\blocgfi{\begin{picture}(80,50)
  \put(60,40){\makebox(0,0){$\rf_i$}}
  \put(60,40){\circle{20}}
  \put(60,30){\vector(0,-1){20}}
  \put(60,0){\circle{20}}
  \put(60,0){\makebox(0,0){$\blue{g}_i$}}
  \put(20,0){\circle{20}}
  \put(20,0){\makebox(0,0){$\epsilon_i$}}
  \put(30,0){\vector(1,0){20}}
\end{picture}
}
\def\blocghfi{\begin{picture}(80,50)
  \put(60,40){\makebox(0,0){$\rf_i$}}
  \put(60,40){\circle{20}}
  \put(60,30){\vector(0,-1){20}}
  \put(65,20){\makebox(0,0){$h$}}
  \put(60,0){\circle{20}}
  \put(60,0){\makebox(0,0){$\blue{g}_i$}}
  \put(20,0){\circle{20}}
  \put(20,0){\makebox(0,0){$\epsilon_i$}}
  \put(30,0){\vector(1,0){20}}
\end{picture}
}
\def\blocgie{\begin{picture}(80,50)
  \put(60,40){\makebox(0,0){$\red{m}$}}
  \put(60,40){\circle{20}}
  \put(60,30){\vector(0,-1){20}}
  \put(60,0){\circle{20}}
  \put(60,0){\makebox(0,0){$\blue{g}_i$}}
  \put(30,0){\circle{20}}
  \put(30,0){\makebox(0,0){$\epsilon_i$}}
  \put(40,0){\vector(1,0){10}}
  \put(0,0){\circle{20}}
  \put(0,0){\makebox(0,0){$\rve$}}
  \put(10,0){\vector(1,0){10}}
\end{picture}
}
\def\blocgievei{\begin{picture}(80,50)
  \put(60,40){\makebox(0,0){$\red{m}$}}
  \put(60,40){\circle{20}}
  \put(60,30){\vector(0,-1){20}}
  \put(60,0){\circle{20}}
  \put(60,0){\makebox(0,0){$\blue{g}_i$}}
  \put(30,0){\circle{20}}
  \put(30,0){\makebox(0,0){$\epsilon_i$}}
  \put(40,0){\vector(1,0){10}}
  \put(0,0){\circle{20}}
  \put(0,0){\makebox(0,0){$\rvei$}}
  \put(10,0){\vector(1,0){10}}
\end{picture}
}
\def\blocfgie{\begin{picture}(80,100)
  \put(60,40){\makebox(0,0){$\rf_i$}}
  \put(60,40){\circle{20}}
  \put(60,30){\vector(0,-1){20}}
  \put(65,20){\makebox(0,0){$\blue{h}$}}
  
  \put(60,0){\circle{20}}
  \put(60,0){\makebox(0,0){$\blue{g}_i$}}
  \put(30,0){\circle{20}}
  \put(30,0){\makebox(0,0){$\epsilon_i$}}
  \put(40,0){\vector(1,0){10}}
  \put(0,0){\circle{20}}
  \put(0,0){\makebox(0,0){$\rve$}}
  \put(10,0){\vector(1,0){10}}
\end{picture}
}
\def\blocvfgie{\begin{picture}(80,100)
  \put(60,80){\makebox(0,0){$\rv_i$}}
  \put(60,80){\circle{20}}
  \put(60,70){\vector(0,-1){20}}
  
  \put(60,40){\makebox(0,0){$\rf_i$}}
  \put(60,40){\circle{20}}
  \put(60,30){\vector(0,-1){20}}
  \put(65,20){\makebox(0,0){$\blue{h}$}}
  
  \put(60,0){\circle{20}}
  \put(60,0){\makebox(0,0){$\blue{g}_i$}}
  \put(30,0){\circle{20}}
  \put(30,0){\makebox(0,0){$\epsilon_i$}}
  \put(40,0){\vector(1,0){10}}
  \put(0,0){\circle{20}}
  \put(0,0){\makebox(0,0){$\rve$}}
  \put(10,0){\vector(1,0){10}}
\end{picture}
}
\def\blocvfhgie{\begin{picture}(80,100)
  \put(60,80){\makebox(0,0){$\rv_i$}}
  \put(60,80){\circle{20}}
  \put(60,70){\vector(0,-1){20}}
  
  \put(60,40){\makebox(0,0){$\rf_i$}}
  \put(60,40){\circle{20}}
  \put(60,30){\vector(0,-1){20}}
  
  \put(30,30){\circle{20}}
  \put(30,30){\makebox(0,0){$\red{h}$}}
  \put(38,22){\vector(1,-1){15}}
  
  \put(60,0){\circle{20}}
  \put(60,0){\makebox(0,0){$\blue{g}_i$}}
  \put(30,0){\circle{20}}
  \put(30,0){\makebox(0,0){$\epsilon_i$}}
  \put(40,0){\vector(1,0){10}}
  \put(0,0){\circle{20}}
  \put(0,0){\makebox(0,0){$\rve$}}
  \put(10,0){\vector(1,0){10}}
\end{picture}
}
\def\blocvfhgiei{\begin{picture}(80,100)
  \put(60,80){\makebox(0,0){$\rv_i$}}
  \put(60,80){\circle{20}}
  \put(60,70){\vector(0,-1){20}}
  
  \put(60,40){\makebox(0,0){$\rf_i$}}
  \put(60,40){\circle{20}}
  \put(60,30){\vector(0,-1){20}}
  
  \put(30,30){\circle{20}}
  \put(30,30){\makebox(0,0){$\red{h}$}}
  \put(38,22){\vector(1,-1){15}}
  
  \put(60,0){\circle{20}}
  \put(60,0){\makebox(0,0){$\blue{g}_i$}}
  \put(30,0){\circle{20}}
  \put(30,0){\makebox(0,0){$\epsilon_i$}}
  \put(40,0){\vector(1,0){10}}
  \put(0,0){\circle{20}}
  \put(0,0){\makebox(0,0){$\rvei$}}
  \put(10,0){\vector(1,0){10}}
\end{picture}
}
\def\blocfgHe{\begin{picture}(80,100)
  \put(60,40){\makebox(0,0){$\rfb$}}
  \put(60,40){\circle{20}}
  \put(60,30){\vector(0,-1){20}}
  \put(65,20){\makebox(0,0){$\blue{\Hb}$}}
  
  \put(60,0){\circle{20}}
  \put(60,0){\makebox(0,0){$\bgb$}}
  \put(30,0){\circle{20}}
  \put(30,0){\makebox(0,0){$\epsilonb$}}
  \put(40,0){\vector(1,0){10}}
  \put(0,0){\circle{20}}
  \put(0,0){\makebox(0,0){$\rve$}}
  \put(10,0){\vector(1,0){10}}
\end{picture}
}
\def\blocvfgHe{\begin{picture}(80,100)
  \put(60,80){\makebox(0,0){$\rvb$}}
  \put(60,80){\circle{20}}
  \put(60,70){\vector(0,-1){20}}
  
  \put(60,40){\makebox(0,0){$\rfb$}}
  \put(60,40){\circle{20}}
  \put(60,30){\vector(0,-1){20}}
  \put(65,20){\makebox(0,0){$\bHb$}}
  
  \put(60,0){\circle{20}}
  \put(60,0){\makebox(0,0){$\bgb$}}
  \put(30,0){\circle{20}}
  \put(30,0){\makebox(0,0){$\epsilonb$}}
  \put(40,0){\vector(1,0){10}}
  \put(0,0){\circle{20}}
  \put(0,0){\makebox(0,0){$\rve$}}
  \put(10,0){\vector(1,0){10}}
\end{picture}
}
\def\blocvfgHei{\begin{picture}(80,100)
  \put(60,80){\makebox(0,0){$\rvb$}}
  \put(60,80){\circle{20}}
  \put(60,70){\vector(0,-1){20}}
  
  \put(60,40){\makebox(0,0){$\rfb$}}
  \put(60,40){\circle{20}}
  \put(60,30){\vector(0,-1){20}}
  \put(65,20){\makebox(0,0){$\bHb$}}
  
  \put(60,0){\circle{20}}
  \put(60,0){\makebox(0,0){$\bgb$}}
  \put(30,0){\circle{20}}
  \put(30,0){\makebox(0,0){$\epsilonb$}}
  \put(40,0){\vector(1,0){10}}
  \put(0,0){\circle{20}}
  \put(0,0){\makebox(0,0){$\rvei$}}
  \put(10,0){\vector(1,0){10}}
\end{picture}
}
\def\blocvfHge{\begin{picture}(80,100)
  \put(60,80){\makebox(0,0){$\rvb$}}
  \put(60,80){\circle{20}}
  \put(60,70){\vector(0,-1){20}}
  
  \put(60,40){\makebox(0,0){$\rfb$}}
  \put(60,40){\circle{20}}
  \put(60,30){\vector(0,-1){20}}
  
  \put(30,30){\circle{20}}
  \put(30,30){\makebox(0,0){$\rHb$}}
  \put(38,22){\vector(1,-1){15}}
  
  \put(60,0){\circle{20}}
  \put(60,0){\makebox(0,0){$\bgb$}}
  \put(30,0){\circle{20}}
  \put(30,0){\makebox(0,0){$\epsilonb$}}
  \put(40,0){\vector(1,0){10}}
  \put(0,0){\circle{20}}
  \put(0,0){\makebox(0,0){$\rve$}}
  \put(10,0){\vector(1,0){10}}
\end{picture}
}
\def\blocvfHgei{\begin{picture}(80,100)
  \put(60,80){\makebox(0,0){$\rvb$}}
  \put(60,80){\circle{20}}
  \put(60,70){\vector(0,-1){20}}
  
  \put(60,40){\makebox(0,0){$\rfb$}}
  \put(60,40){\circle{20}}
  \put(60,30){\vector(0,-1){20}}
  
  \put(30,30){\circle{20}}
  \put(30,30){\makebox(0,0){$\rHb$}}
  \put(38,22){\vector(1,-1){15}}
  
  \put(60,0){\circle{20}}
  \put(60,0){\makebox(0,0){$\bgb$}}
  \put(30,0){\circle{20}}
  \put(30,0){\makebox(0,0){$\epsilonb$}}
  \put(40,0){\vector(1,0){10}}
  \put(0,0){\circle{20}}
  \put(0,0){\makebox(0,0){$\rvei$}}
  \put(10,0){\vector(1,0){10}}
\end{picture}
}
\def\blocvhFge{\begin{picture}(80,100)
  \put(60,80){\makebox(0,0){$\rvb$}}
  \put(60,80){\circle{20}}
  \put(60,70){\vector(0,-1){20}}
  
  \put(60,40){\makebox(0,0){$\rhb$}}
  \put(60,40){\circle{20}}
  \put(60,30){\vector(0,-1){20}}
  
  \put(30,30){\circle{20}}
  \put(30,30){\makebox(0,0){$\bFb$}}
  \put(38,22){\vector(1,-1){15}}

  \put(60,0){\circle{20}}
  \put(60,0){\makebox(0,0){$\bgb$}}
  \put(30,0){\circle{20}}
  \put(30,0){\makebox(0,0){$\epsilonb$}}
  \put(40,0){\vector(1,0){10}}
  \put(0,0){\circle{20}}
  \put(0,0){\makebox(0,0){$\rve$}}
  \put(10,0){\vector(1,0){10}}
\end{picture}
}
\def\blocvhFgei{\begin{picture}(80,100)
  \put(60,80){\makebox(0,0){$\rvb$}}
  \put(60,80){\circle{20}}
  \put(60,70){\vector(0,-1){20}}
  
  \put(60,40){\makebox(0,0){$\rhb$}}
  \put(60,40){\circle{20}}
  \put(60,30){\vector(0,-1){20}}
  
  \put(30,30){\circle{20}}
  \put(30,30){\makebox(0,0){$\bFb$}}
  \put(38,22){\vector(1,-1){15}}

  \put(60,0){\circle{20}}
  \put(60,0){\makebox(0,0){$\bgb$}}
  \put(30,0){\circle{20}}
  \put(30,0){\makebox(0,0){$\epsilonb$}}
  \put(40,0){\vector(1,0){10}}
  \put(0,0){\circle{20}}
  \put(0,0){\makebox(0,0){$\rvei$}}
  \put(10,0){\vector(1,0){10}}
\end{picture}
}
\def\blochFge{\begin{picture}(80,40)
  \put(60,40){\makebox(0,0){$\rhb$}}
  \put(60,40){\circle{20}}
  \put(60,30){\vector(0,-1){20}}
  
  \put(30,30){\circle{20}}
  \put(30,30){\makebox(0,0){$\bFb$}}
  \put(38,22){\vector(1,-1){15}}

  \put(60,0){\circle{20}}
  \put(60,0){\makebox(0,0){$\bgb$}}
  \put(30,0){\circle{20}}
  \put(30,0){\makebox(0,0){$\epsilonb$}}
  \put(40,0){\vector(1,0){10}}
  \put(0,0){\circle{20}}
  \put(0,0){\makebox(0,0){$\rve$}}
  \put(10,0){\vector(1,0){10}}
\end{picture}
}
\def\blochFgei{\begin{picture}(80,40)
  \put(60,40){\makebox(0,0){$\rhb$}}
  \put(60,40){\circle{20}}
  \put(60,30){\vector(0,-1){20}}
  
  \put(30,30){\circle{20}}
  \put(30,30){\makebox(0,0){$\bFb$}}
  \put(38,22){\vector(1,-1){15}}

  \put(60,0){\circle{20}}
  \put(60,0){\makebox(0,0){$\bgb$}}
  \put(30,0){\circle{20}}
  \put(30,0){\makebox(0,0){$\epsilonb$}}
  \put(40,0){\vector(1,0){10}}
  \put(0,0){\circle{20}}
  \put(0,0){\makebox(0,0){$\rvei$}}
  \put(10,0){\vector(1,0){10}}
\end{picture}
}
\def\blocfHge{\begin{picture}(80,40)  
  \put(60,40){\makebox(0,0){$\bfb$}}
  \put(60,40){\circle{20}}
  \put(60,30){\vector(0,-1){20}}
  
  \put(30,30){\circle{20}}
  \put(30,30){\makebox(0,0){$\rHb$}}
  \put(38,22){\vector(1,-1){15}}
  
  \put(60,0){\circle{20}}
  \put(60,0){\makebox(0,0){$\bgb$}}
  \put(30,0){\circle{20}}
  \put(30,0){\makebox(0,0){$\epsilonb$}}
  \put(40,0){\vector(1,0){10}}
  \put(0,0){\circle{20}}
  \put(0,0){\makebox(0,0){$\rve$}}
  \put(10,0){\vector(1,0){10}}
\end{picture}
}
\def\blocfHgei{\begin{picture}(80,40)  
  \put(60,40){\makebox(0,0){$\bfb$}}
  \put(60,40){\circle{20}}
  \put(60,30){\vector(0,-1){20}}
  
  \put(30,30){\circle{20}}
  \put(30,30){\makebox(0,0){$\rHb$}}
  \put(38,22){\vector(1,-1){15}}
  
  \put(60,0){\circle{20}}
  \put(60,0){\makebox(0,0){$\bgb$}}
  \put(30,0){\circle{20}}
  \put(30,0){\makebox(0,0){$\epsilonb$}}
  \put(40,0){\vector(1,0){10}}
  \put(0,0){\circle{20}}
  \put(0,0){\makebox(0,0){$\rvei$}}
  \put(10,0){\vector(1,0){10}}
\end{picture}
}
\def\blocFgHe{\begin{picture}(80,40)
  \put(60,40){\makebox(0,0){$\bfb$}}
  \put(60,40){\circle{20}}
  \put(60,30){\vector(0,-1){20}}
  \put(65,20){\makebox(0,0){$\blue{\rHb}$}}
  
  \put(60,0){\circle{20}}
  \put(60,0){\makebox(0,0){$\bgb$}}
  \put(30,0){\circle{20}}
  \put(30,0){\makebox(0,0){$\epsilonb$}}
  \put(40,0){\vector(1,0){10}}
  \put(0,0){\circle{20}}
  \put(0,0){\makebox(0,0){$\rve$}}
  \put(10,0){\vector(1,0){10}}
\end{picture}
}

\def\ftrfi{\bg(r,\phi)}
\def\wth{\widehat{\tilde{h}}}
\def\varx#1#2{\mbox{Var}_{#1}\left\{#2\right\}}
\def\rvbxi{\rvb_{\xi}}
\def\rVbxi{\rVb_{\xi}}
\def\rvbxih{\rvbh_{\xi}}

\def\repsilon{\red{\epsilon}}

\def\vzj{v_{z_j}}
\def\vxij{v_{\xi_j}}
\def\vei{v_{\epsilon_i}}

\def\vzjh{\widehat{v}_{z_j}}
\def\vxijh{\widehat{v}_{\xi_j}}
\def\veih{\widehat{v}_{\epsilon_i}}

\def\vbeh{\widehat{\vb}_{\epsilon}}
\def\vbxih{\widehat{\vb}_{\xi}}
\def\vbzh{\widehat{\vb}_{z}}

\def\Vbeh{\widehat{\Vb}_{\epsilon}}
\def\Vbxih{\widehat{\Vb}_{\xi}}
\def\Vbzh{\widehat{\Vb}_{z}}

\def\Ybeh{\widehat{\Yb}_{\epsilon}}
\def\Ybxih{\widehat{\Yb}_{\xi}}
\def\Ybzh{\widehat{\Yb}_{z}}

\def\vf{v_f}
\def\vbe{\vb_{\epsilon}}
\def\vbz{\vb_z}
\def\vxi{v_{\xi}}
\def\vbxi{\vb_{\xi}}
\def\Vbe{\Vb_{\epsilon}}
\def\Vbxi{\Vb_{\xi}}
\def\Vbz{\Vb_z}

\def\half{\frac{1}{2}}

\def\dpdx#1#2{\frac{\partial {#1}}{\partial {#2}}}
\def\dpdxd#1#2{\frac{\partial^2 {#1}}{\partial {#2}^2}}
\def\dpdxdy#1#2#3{{{\partial ^2 {#1}}\over{\partial {#2} \partial {#3}}}}

\def\KL#1#2{\mbox{KL}\left[#1 : #2\right]}
\def\Covx#1#2{\mbox{Cov}_{#1}(#2)}
\def\braket#1#2{<{#1}(#2)>}

\def\rlambdab{\red{\lambdab}}
\def\bve{\blue{\ve}}
\def\ralphafjt{\red{\tilde{\alpha}_{f_j}}}
\def\rbetafjt{\red{\tilde{\beta}_{f_j}}}
\def\rSigmabt{\red{\tilde{\Sigmab}}}
\def\rfjt{\red{\tilde{f_j}}}
\def\zetab{\bm{\zeta}}
\def\bvxi{\blue{v_\xi}}
\def\bgamma{\blue{\gamma}}
\def\rgamma{\red{\gamma}}
\def\alphagamz{\alpha_{\gamma_0}}
\def\betagamz{\beta_{\gamma_0}}
\def\alphazetaz{\alpha_{\zeta_0}}
\def\betazetaz{\beta_{\zeta_0}}

\def\bvf{\blue{v_f}}
\def\bvzeta{\blue{v_\zeta}}
\def\rgb{\red{\gb}}
\def\rvxii{\red{v_{\xi_i}}}
\def\rvzeta{\red{v_{\zeta}}}

\usepackage{tikz,mathpazo}
\usetikzlibrary{shapes.geometric, arrows}
\usetikzlibrary{positioning,shapes,shadows,arrows}
\usetikzlibrary{mindmap,backgrounds}
\tikzstyle{startstop} = [rectangle, rounded corners, minimum width=1.5cm, minimum height=0.5cm,text centered, draw=black]
\tikzstyle{io} = [trapezium, trapezium left angle=70, trapezium right angle=110, minimum width=1.5cm, minimum height=0.6cm, text centered, draw=black]
\tikzstyle{process} = [rectangle, minimum width=1cm, minimum height=0.5cm, text centered, draw=black]
\tikzstyle{note} = [text centered]
\tikzstyle{decision} = [diamond, minimum width=1.5cm, minimum height=0.3cm, text centered, draw=black]
\tikzstyle{arrow} = [thick,->,>=stealth]

\def\blambda{\blue{\lambda}}
\def\blambdab{\blue{\lambdab}}

\def\rWb{\red{\Wb}}
\def\rWbh{\red{\widehat{\Wb}}}